\documentclass[letterpaper]{article}
\usepackage[preprint]{aaai2027}
\usepackage[hyphens]{url}  % DO NOT CHANGE THIS
\usepackage{graphicx} % DO NOT CHANGE THIS
\usepackage{natbib}  % DO NOT CHANGE THIS AND DO NOT ADD ANY OPTIONS TO IT
\usepackage{caption} % DO NOT CHANGE THIS AND DO NOT ADD ANY OPTIONS TO IT
\usepackage{booktabs}
\usepackage{multirow}
\usepackage{amsmath}
\usepackage{amssymb}
\usepackage{placeins}
\usepackage{tabularx}
\usepackage{xcolor}

\title{CAPE-T2V: Captioner-Anchored Prompt Enhancement toward Two-Sided Conditioning Alignment in Text-to-Video Generation}
\author{Yizhuo Jia\textsuperscript{1,2}\thanks{These authors contributed equally.}\thanks{This work was conducted during the author's internship at Kling Team.},
Jingyun Hua\textsuperscript{2}\footnotemark[1]\thanks{Corresponding author.},
Yuanxing Zhang\textsuperscript{2}}
\affiliations{\textsuperscript{1}Fudan University \\
\textsuperscript{2}Kling Team}

\begin{document}
\maketitle

\begin{abstract}
Text-to-video (T2V) diffusion transformers (DiTs) are trained with detailed video captions, whereas inference often relies on user prompts rewritten by a prompt enhancer (PE). Prior work has improved generation by optimizing the PE, the DiT, or both; some methods have also sought to narrow the training--inference mismatch through shared schemas. Yet even within a shared schema, inference-time PE outputs and DiT training captions may still differ in detail selection, information organization, descriptive granularity, and phrasing. We refer to this residual mismatch as the \emph{PE--Caption gap} and introduce \textbf{CAPE-T2V}, a two-step Captioner-Anchored Prompt Enhancement framework toward two-sided conditioning alignment in T2V generation. First, for each captioner-generated target, CAPE-T2V constructs three PE training examples using a concise source caption, a detailed source caption, and a pseudo user prompt derived from that target. It then fine-tunes the PE to map each input to the target. Second, CAPE-T2V fine-tunes the DiT on video-derived captions rewritten by the Anchored PE; the same PE rewrites user prompts at inference. Relative to a baseline using the same caption schema, CAPE-T2V achieves higher aggregate scores on StoryEval, VBench-2.0, and T2V-CompBench across Wan2.2 and LTX-2.3. Further, CAPE-T2V exhibits a smaller PE--Caption gap than the baseline: its DiT fine-tuning captions are closer in distribution to inference-time PE outputs, as measured by embedding-based $\mathrm{MMD}^2$. Overall, these results support CAPE-T2V as an effective approach to mitigating the PE--Caption gap. The project is available at \url{https://github.com/yizzz927/CAPE-T2V}.
\end{abstract}

\section{Introduction}
\begin{figure}[!t]
\centering
\includegraphics[width=\columnwidth]{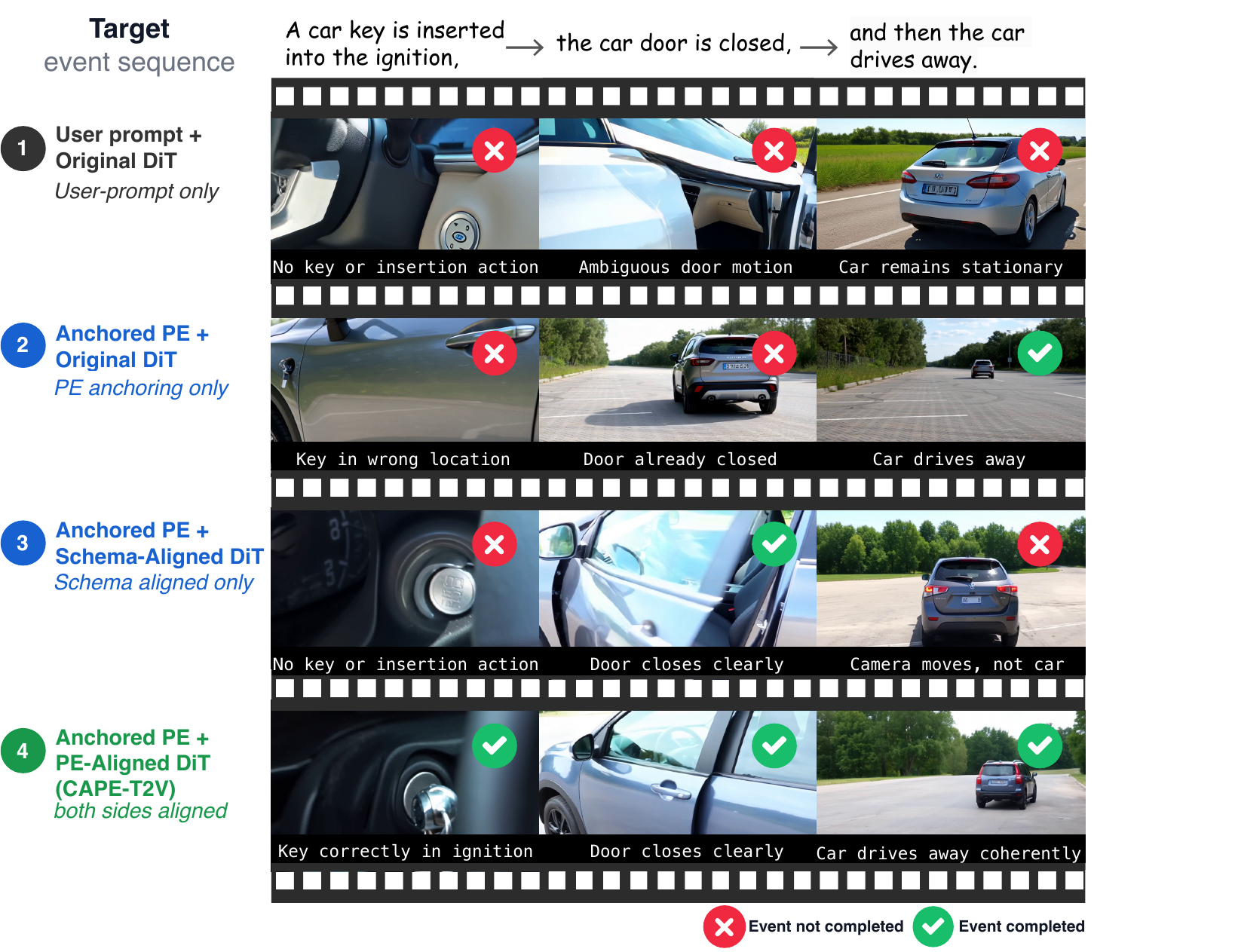}
\caption{\textbf{Partial alignment is not enough.} A StoryEval prompt requests three consecutive events. Conditioning the Original DiT directly on the user prompt completes none of them. The Anchored PE with the Original DiT and with the Schema-Aligned DiT each recovers a different single event. Only CAPE-T2V---where the same Anchored PE both writes the DiT's fine-tuning captions and rewrites the user prompt at inference---completes all three. Per-event verdicts use the same three-call unanimous GPT-5.5 protocol as our StoryEval evaluation.}
\label{fig:case_car_key}
\end{figure}

Recent advances in diffusion transformers (DiTs) have improved text-to-video (T2V) generation~\citep{wan22,ltx2,hunyuanvideo15}. These models learn text--video correspondence from videos paired with detailed captions, whereas users write short prompts. Deployed systems therefore place a prompt enhancer (PE) in front of the generator to rewrite user prompts into richer conditioning text. Once deployed, the PE becomes part of the model's conditioning interface: its outputs are the text that actually conditions the DiT. The DiT is thus trained on detailed video captions but conditioned on PE outputs at inference, so generation quality can depend on how compatible these two text distributions are.

Prior work on prompt-conditioned visual generation intervenes at the prompt, the generator, or both. Prompt-side methods include training-free optimization of the text prompt and initial diffusion noise~\citep{pos}, as well as learned prompt enhancers trained with supervised targets and feedback associated with generated outputs~\citep{promptist,vpo,promptavideo}. VideoDPO~\citep{videodpo} adapts a pretrained video diffusion model using preference pairs, while PromptRL~\citep{promptrl} jointly optimizes a prompt enhancer and generator in a reinforcement learning loop. Prompt- or generator-only approaches target prompt or generation quality but constrain the relation between training-side captions and inference-time PE outputs only indirectly. Joint methods such as PromptRL can coordinate both components more directly, but tie the prompt enhancer to a particular generator.

A complementary line explicitly reduces the train--inference text mismatch by recaptioning training data toward expected prompts, fitting a PE to training descriptions, or imposing a shared caption structure~\citep{recap,promptcot,instancecap}. The first two derive their target text from training data, whereas the third prescribes a schema. For a deployed generator, its original caption sources and captioning recipe may be unavailable. Moreover, a shared schema fixes only which parts appear and in what order; it does not determine within-schema realization choices such as detail selection, information organization, descriptive granularity, or phrasing. Thus, even under a shared schema, DiT fine-tuning captions can remain distributionally different from the inference-time outputs produced by the deployed PE. We call this residual distributional discrepancy the \emph{PE--Caption gap}. Figure~\ref{fig:case_car_key} presents a selected qualitative case in which the matched Schema-Aligned control and CAPE-T2V yield different event-completion outcomes under the same schema.

We address this residual with \textbf{CAPE-T2V}, a two-step Captioner-Anchored Prompt Enhancement framework that does not require the target generator's native caption source or captioning recipe. Rather than imitating a generator-specific caption distribution, we use an external captioner to produce PE training targets under a selected schema.

CAPE-T2V implements this idea in two steps. Step~1 anchors the PE to captioner-generated targets, yielding the Captioner-Anchored PE (Anchored PE), which is frozen thereafter. In Step~2, the Anchored PE rewrites video-derived dense captions, and the resulting captions are used to fine-tune the PE-Aligned DiT. Pairing this DiT with the same Anchored PE, which rewrites user prompts at inference, forms the complete CAPE-T2V system. This construction requires no joint PE--DiT optimization.

To evaluate the benefit of addressing the residual PE--Caption gap beyond schema matching alone, we compare CAPE-T2V with a matched schema-aligned control, termed the Schema-Aligned DiT. Both systems use the same videos, dense-caption sources, caption schema, DiT optimization, training budget, and Anchored PE at inference. They differ only in how the DiT fine-tuning captions are constructed: the control uses a separately prompted schema-aligned rewriter, whereas CAPE-T2V uses the Anchored PE. CAPE-T2V achieves higher aggregate scores in all six model--benchmark pairs formed by the two T2V models and three benchmarks. Its gains on StoryEval, VBench-2.0, and T2V-CompBench are 1.6, 1.12, and 0.30 percentage points for Wan2.2, and 1.4, 0.92, and 0.65 percentage points for LTX-2.3, respectively. In the token-length-matched analysis, the DiT fine-tuning captions produced by the Anchored PE have lower embedding-based MMD$^2$ to fixed-seed inference-time Anchored PE rewrites of the benchmark user prompts than those produced by the prompted rewriter (0.0682 versus 0.0764), indicating a smaller measured embedding discrepancy.

In summary, our contributions are:
\begin{itemize}
\item \textbf{The PE--Caption gap beyond schema matching.} We identify a residual conditioning mismatch that persists even when inference-time PE outputs and DiT fine-tuning captions share a schema. We illustrate possible within-schema realization differences and quantify the resulting overall distributional discrepancy in a fixed embedding space using MMD$^2$.

\item \textbf{Captioner anchoring toward two-sided conditioning alignment.} CAPE-T2V anchors the PE using diverse inputs paired with captioner-generated targets under the selected schema. It then uses the Anchored PE both to construct DiT fine-tuning captions and to rewrite user prompts at inference, reducing the measured discrepancy between the two conditioning sides without joint PE--DiT optimization.

\item \textbf{Cross-model validation without native caption recipes.} The caption schema can be specified without recovering each target generator's native captioning recipe. Using one shared external schema and the same Anchored PE, CAPE-T2V achieves higher aggregate scores than the Schema-Aligned DiT on each of the three benchmarks for both Wan2.2 and LTX-2.3.
\end{itemize}

\section{Related Work}
\begin{figure*}[!t]
\centering
\includegraphics[width=0.85\textwidth]{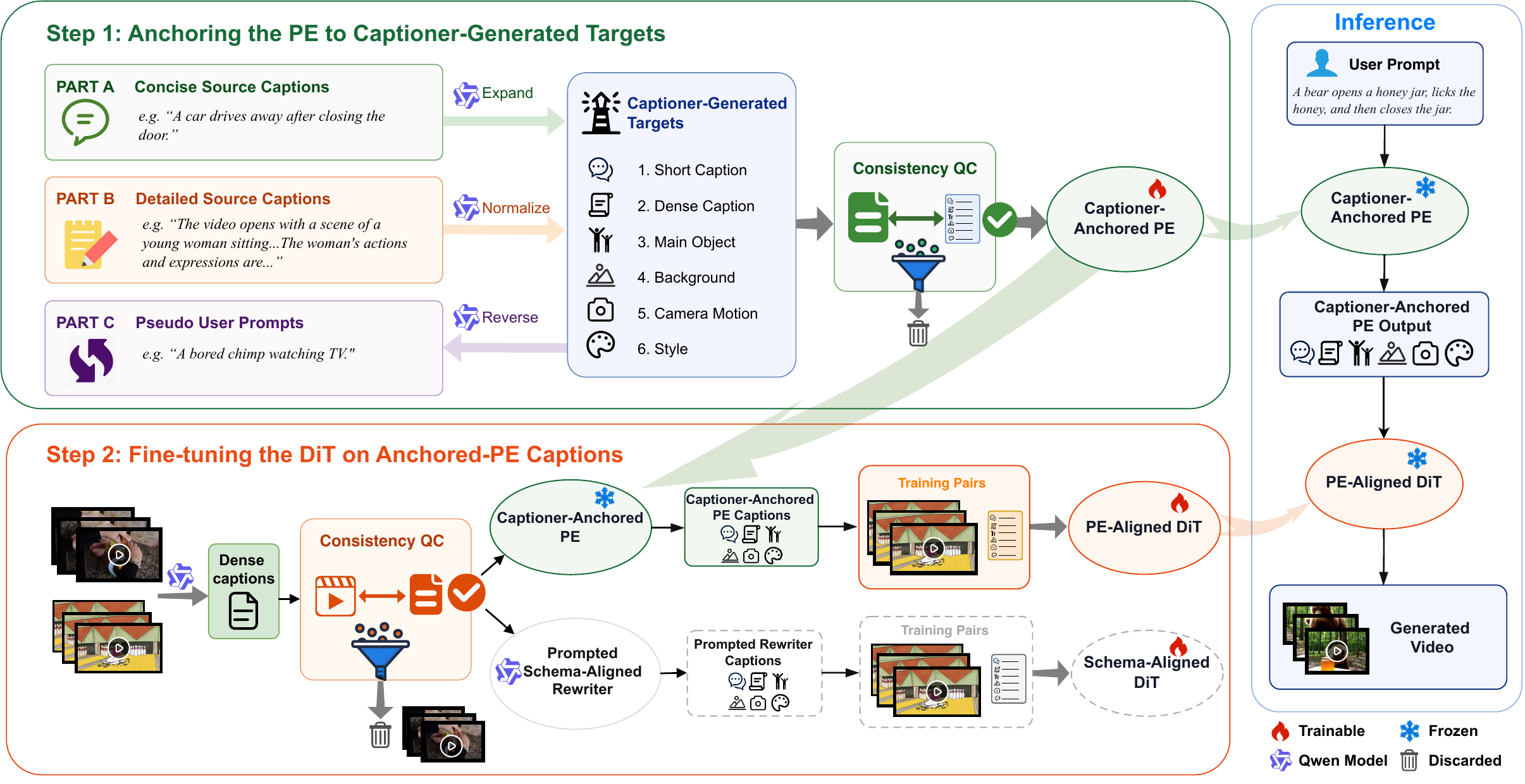}
\caption{\textbf{CAPE-T2V overview and matched comparison.} In \textbf{Step~1}, the PE is trained on concise source captions, detailed source captions, and pseudo user prompts paired with captioner-generated targets, yielding the Anchored PE. In \textbf{Step~2}, captions produced by the Anchored PE are used to fine-tune the PE-Aligned DiT. For the matched comparison, the Schema-Aligned DiT is instead fine-tuned on captions produced by the prompted schema-aligned rewriter; both DiTs use the Anchored PE at inference.}
\label{fig:cape_pipeline}
\end{figure*}
\paragraph{Optimizing the prompt enhancer or the generator.}
Prompt-side methods improve generation without modifying the base generator. POS~\citep{pos} performs training-free optimization of the text prompt and initial diffusion noise, while RAPO and RAPO++~\citep{rapo,rapopp} retrieve and align prompt modifiers. Promptist, VPO, and Prompt-A-Video~\citep{promptist,vpo,promptavideo} instead learn prompt enhancers through combinations of supervised training and reward or preference feedback evaluated on prompts or generated outputs. Recaptioning and caption optimization improve the text supervision available for visual generation~\citep{miradata,vc4vg,avcdpo}. VideoDPO~\citep{videodpo} adapts a pretrained video diffusion model using preference pairs, whereas PromptRL~\citep{promptrl} jointly optimizes a prompt enhancer and an image generator in a reinforcement learning loop, directly coordinating the two components but coupling the prompt enhancer to the target generator.

\paragraph{Aligning training and inference text.}
A complementary line explicitly reduces the mismatch between training and inference text. RECAP~\citep{recap} recaptions training data and shows that the resulting captions reduce train--inference skew while providing richer textual supervision. DALL-E~3~\citep{dalle3} and Movie Gen~\citep{moviegen} combine recaptioned training data with inference-time prompt rewriting, while PromptCoT~\citep{promptcot} adapts a PE toward descriptions of high-quality visual content derived from training data. The latter methods couple the two conditioning distributions in the direction \emph{captions}~$\rightarrow$~\emph{PE}: the training-caption side defines the target form, and the inference-time PE is fitted toward it. This direction commonly relies on access to the relevant training captions or captioning pipeline, which may be unavailable for a deployed generator.

\paragraph{Constraining training and inference text to one interface.}
Closest to our T2V setting is InstanceCap~\citep{instancecap}, which gives training captions and prompt enhancement a shared structure, although different procedures produce the two sets of captions. A shared structure does not determine how content is realized within its parts. We therefore compare captions produced from the same dense-caption sources by different caption operators under one schema. CAPE-T2V couples the two conditioning distributions in the reverse direction, \emph{PE}~$\rightarrow$~\emph{captions}: Step~1 anchors the PE to captioner-generated targets, and Step~2 uses the resulting Anchored PE to produce the DiT training captions; the same PE rewrites user prompts at inference. Our controlled comparison tests whether using the deployed PE to construct DiT fine-tuning captions provides benefits beyond sharing the schema alone.

\section{Method: CAPE-T2V}

CAPE-T2V comprises two sequential steps, summarized in Figure~\ref{fig:cape_pipeline}. Step~1, \emph{Anchoring the PE to Captioner-Generated Targets}, trains the PE to map diverse inputs to captioner-generated targets under the selected schema, yielding the Anchored PE. Step~2, \emph{Fine-tuning the DiT on Anchored PE Captions}, uses the Anchored PE to construct the DiT fine-tuning captions and updates only the DiT.

\paragraph{Problem setup.}
CAPE-T2V targets a mismatch at the DiT conditioning interface. During fine-tuning, a T2V DiT learns from captions paired with training videos, whereas at deployment it is conditioned on PE rewrites of user prompts. To compare alignment with the deployed PE against schema alignment alone, we construct two DiT fine-tuning caption sets from the same filtered dense captions, with both following the same schema. The Anchored PE produces captions used to fine-tune the PE-Aligned DiT, whereas a separately prompted schema-aligned rewriter (hereafter, prompted rewriter) produces captions used to fine-tune the Schema-Aligned DiT. At inference, both DiTs use rewrites produced by the Anchored PE. Thus, both settings are schema aligned, but only the PE-Aligned DiT is trained on captions produced by the deployed PE.

We formalize these paths through three empirical text distributions. $P_{\mathrm{PE}}^{\mathrm{user}}$ denotes Anchored PE outputs from user prompts and represents the inference-time conditions. Given video-derived dense captions, $P_{\mathrm{PE}}^{\mathrm{dense}}$ and $P_{\mathrm{rewriter}}^{\mathrm{dense}}$ denote outputs from the Anchored PE and the prompted rewriter, respectively.

\paragraph{Residual PE--Caption gap.}
Although all compared captions follow the same schema, the schema specifies only the parts and their order, not how content is realized within them. The prompted rewriter and the Anchored PE can therefore induce different within-schema text distributions. We hypothesize that constructing the DiT fine-tuning captions with the deployed PE leaves a smaller gap to the inference-time conditions than schema matching alone:
\[
D\!\left(P_{\mathrm{PE}}^{\mathrm{user}},P_{\mathrm{PE}}^{\mathrm{dense}}\right)
<
D\!\left(P_{\mathrm{PE}}^{\mathrm{user}},P_{\mathrm{rewriter}}^{\mathrm{dense}}\right).
\]
Here, $D$ denotes a discrepancy between text distributions; Section~\ref{sec:gap_analysis} instantiates it as MMD$^2$ in a fixed text-embedding space.

\subsection{Step~1: Anchoring the PE to Captioner-Generated Targets}

We train the PE to map inputs with different levels of detail to captioner-generated targets. This Anchored PE is used twice in our pipeline, and the two uses demand different behavior: expanding concise user prompts at inference, and conservatively normalizing video-derived dense captions for DiT fine-tuning.

\paragraph{Captioner-generated targets.}
We first define the target captions used to train the PE. Recovering the native captioning recipe of every target DiT would make the method model-specific and is often infeasible because such recipes are not fully available. The shared schema may follow the caption schema used to train a target DiT when it is available, or it may be selected according to the application. In our implementation, we organize the six caption types defined by MiraData~\citep{miradata} into a fixed schema: Short Caption, Dense Caption, Main Object Caption, Background Caption, Camera Caption, and Style Caption. These six parts jointly describe global semantics, temporal detail, subjects, scene context, camera behavior, and visual style. This information is useful for T2V conditioning.

After consistency filtering, each retained captioner output becomes a captioner-generated target $y_i$. These targets specify not only which parts are present, but also concrete realizations of their content. This distinction matters because a shared schema does not determine those realizations. Once selected, the schema is used consistently for PE anchoring and DiT caption construction.

\paragraph{Training-pair construction.}
To cover both uses, we pair the same targets with inputs at different levels of completeness. A concise source caption is paired with $y_i$ to teach structured expansion from a compact semantic description. A detailed source caption is paired with the same target to teach conservative normalization. These two constructions cover concise and detailed caption inputs, but neither reflects how users typically formulate requests. We therefore generate a pseudo user prompt from $y_i$, using prompts from VidProM~\citep{vidprom} as few-shot examples of natural user expression, and pair it with the target. This third construction adds user-style inputs without relying on naturally paired user prompts and video-derived captions, which are generally unavailable. All three constructions share the same targets, so the PE learns to emit one realization of the schema regardless of how its input is written.

For notational simplicity, let $x_i$ denote an input from any of the three constructions above and $y_i$ its paired target. With PE parameters $\phi$, the standard autoregressive SFT objective is
\begin{equation}
\mathcal{L}_{\mathrm{SFT}}(\phi)
=-\frac{1}{\sum_i |y_i|}\sum_i\sum_{j=1}^{|y_i|}
\log p_{\phi}\!\left(y_{i,j}\mid x_i,y_{i,<j}\right).
\label{eq:sft_loss}
\end{equation}
Here, $p_\phi$ is the PE's conditional next-token distribution, $i$ indexes training pairs, and $j$ indexes target tokens. The loss covers target tokens only, with input tokens masked, and is normalized by their total count.

\subsection{Step~2: Fine-tuning the DiT on Anchored PE Captions}
\begin{figure*}[!t]
\centering
\includegraphics[width=1.0\textwidth]{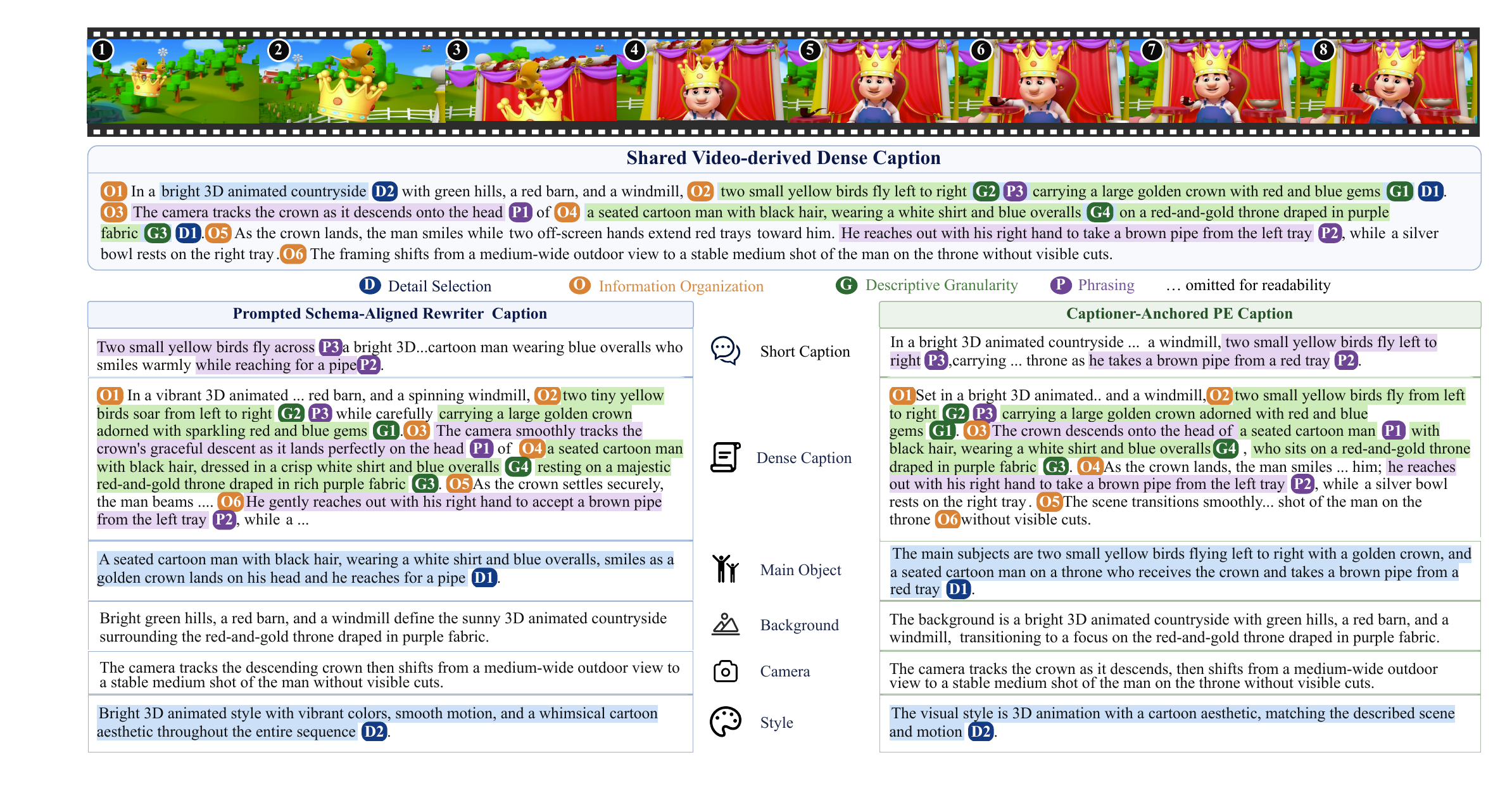}
\caption{\textbf{Four dimensions of residual conditioning mismatch within a shared schema.} Starting from the same video-derived dense caption, the prompted rewriter produces the caption for the Schema-Aligned DiT, whereas the Anchored PE produces the caption for the PE-Aligned DiT. The paired example shows how the two caption operators can differ in detail selection, information organization, descriptive granularity, and phrasing despite following the same six-part schema.}
\label{fig:caption_diff_example}
\end{figure*}

Each training video is paired with a filtered, video-derived dense caption that serves as the shared content source. The Anchored PE used at inference rewrites the dense caption, and its complete six-part output is paired with the source video as the conditioning text for fine-tuning the PE-Aligned DiT. For the controlled baseline, the prompted rewriter processes the same dense caption under the same schema, and its complete six-part output is used to fine-tune the Schema-Aligned DiT. The two paths differ only in which caption operator produces the DiT fine-tuning captions. Figure~\ref{fig:caption_diff_example} shows how their outputs can differ within the shared schema.

Both fine-tuned DiTs retain their own flow-matching formulation, with the schedule $\sigma_t$ and the weighting $w(t)$ taken from each model's own scheduler. Let $z_0$ be the clean video latent, $c$ the conditioning text, $\theta$ the DiT parameters, and $\epsilon\sim\mathcal{N}(0,I)$. At timestep $t$, $z_t=(1-\sigma_t)z_0+\sigma_t\epsilon$ and the velocity target is $v_t=\epsilon-z_0$. We optimize
\begin{equation}
\mathcal{L}_{\mathrm{FM}}(\theta)
=\mathbb{E}_{(z_0,c),\,t,\,\epsilon}\!\left[
w(t)\,\operatorname{mean}\!\left[
\left(v_\theta(z_t,t,c)-v_t\right)^2
\right]
\right].
\label{eq:fm_loss}
\end{equation}
Here, $v_\theta$ is the predicted velocity and the mean is taken over all video-latent elements. CAPE-T2V changes the conditioning text $c$, not the DiT architecture or training objective, and introduces no auxiliary loss.
% \paragraph{Inference.}

At inference, the Anchored PE rewrites each user prompt, and the complete rewrite conditions the PE-Aligned DiT. Section~\ref{sec:gap_analysis} measures the discrepancy between the corresponding text distributions.

\section{Experiments}

\subsection{Experimental Setup}
\paragraph{Benchmarks and evaluation.}
We evaluate CAPE-T2V with two T2V models, \textbf{Wan2.2-T2V-A14B}~\citep{wan22,wan22model} and \textbf{LTX-2.3}~\citep{ltx2,ltx23model}, on StoryEval~\citep{storyeval}, VBench-2.0~\citep{vbench2}, and T2V-CompBench~\citep{t2vcompbench}. StoryEval evaluates structured multi-event generation. VBench-2.0 evaluates intrinsic faithfulness across human fidelity, controllability, creativity, physics, and commonsense, while T2V-CompBench evaluates compositional generation involving attributes, relationships, interactions, and numeracy. We use the official metrics and aggregation code for VBench-2.0 and T2V-CompBench. For StoryEval, we follow the official prompt set and its three-call unanimous event-completion criterion, substituting GPT-5.5 as the judge. For readability, Table~\ref{tab:main_results} reports the aggregate VBench-2.0 and T2V-CompBench scores; results for individual dimensions and categories are provided in the appendix.
\begin{table*}[t]
\centering
\footnotesize
\setlength{\tabcolsep}{2.1pt}
\renewcommand{\arraystretch}{1.08}
\begin{tabular*}{\textwidth}{@{\extracolsep{\fill}}cccccccccccc@{}}
\toprule
\multirow{3}{*}{\textbf{PE}} &
\multirow{3}{*}{\textbf{DiT}} &
\multicolumn{8}{c}{\textbf{StoryEval}} &
\multirow{3}{*}{\shortstack{\textbf{VBench-2.0}\\\textbf{Score}$\uparrow$}} &
\multirow{3}{*}{\shortstack{\textbf{T2V-CompBench}\\\textbf{Score}$\uparrow$}} \\
\cmidrule(lr){3-10}
& &
\multicolumn{5}{c}{\textbf{Semantic category}} &
\multicolumn{2}{c}{\textbf{Difficulty}} &
\multirow{2}{*}{\textbf{Overall}$\uparrow$} & & \\
\cmidrule(lr){3-7}\cmidrule(lr){8-9}
& & \textbf{Human} & \textbf{Animal} & \textbf{Object} & \textbf{Retrieval} & \textbf{Creative} & \textbf{Easy} & \textbf{Hard} & & & \\
\midrule
\multicolumn{12}{l}{\textbf{Wan2.2}} \\
None & Original DiT & 43.7 & 48.1 & 43.1 & 59.1 & 34.5 & 63.3 & 32.8 & 45.9 & 55.01 & 55.90 \\
Official PE & Original DiT & 48.1 & 49.4 & 44.3 & 59.7 & 37.4 & 66.2 & 33.0 & 47.5 & 55.78 & 59.34 \\
Anchored PE & Original DiT & \underline{66.8} & \underline{66.8} & 60.5 & 68.6 & 54.5 & \underline{81.1} & 45.3 & 64.6 & 60.87 & 65.74 \\
Anchored PE & Schema-Aligned DiT & \textbf{74.5} & 65.4 & \underline{64.5} & \underline{76.6} & \underline{57.7} & 79.9 & \textbf{54.4} & \underline{67.6} & \underline{61.73} & \underline{66.36} \\
Anchored PE & PE-Aligned DiT & \textbf{74.5} & \textbf{69.5} & \textbf{67.5} & \textbf{77.0} & \textbf{60.5} & \textbf{81.3} & \underline{51.8} & \textbf{69.2} & \textbf{62.85} & \textbf{66.66} \\
\midrule
\multicolumn{12}{l}{\textbf{LTX-2.3}} \\
None & Original DiT & 30.0 & 30.5 & 25.5 & 49.7 & 19.4 & 46.8 & 12.6 & 29.1 & 52.88 & 42.43 \\
Official PE & Original DiT & 55.1 & 56.6 & \textbf{47.2} & \textbf{65.7} & 44.6 & \textbf{73.1} & 38.4 & 53.7 & \underline{60.74} & 57.13 \\
Anchored PE & Original DiT & 56.7 & 56.0 & 46.3 & 60.1 & 41.2 & \underline{72.2} & \textbf{39.4} & 53.1 & 60.50 & 58.28 \\
Anchored PE & Schema-Aligned DiT & \underline{57.8} & \underline{57.4} & \textbf{47.2} & 54.0 & \textbf{45.5} & 71.0 & 38.8 & \underline{54.0} & 60.63 & \underline{59.90} \\
Anchored PE & PE-Aligned DiT & \textbf{61.2} & \textbf{58.8} & \underline{46.8} & \underline{61.6} & \underline{45.1} & 72.1 & \underline{39.2} & \textbf{55.4} & \textbf{61.55} & \textbf{60.55} \\
\bottomrule
\end{tabular*}
\caption{Main benchmark results (\%). StoryEval columns report event-completion rates, with Overall aggregated over the full prompt set. Pairing the Anchored PE with the PE-Aligned DiT gives CAPE-T2V. Higher is better; bold and underline mark the best and second-best results within each model family.}
\label{tab:main_results}
\end{table*}
\paragraph{Training data and implementation.}
PE fine-tuning uses three types of input--target pairs constructed with \textbf{Qwen3.5-397B-A17B}~\citep{qwen35vl}. The model generates captioner targets for concise and detailed source captions and then generates pseudo user prompts from those targets. After pairing each input with its corresponding target, filtering the resulting pairs for consistency, and reserving a validation split, approximately 735K pairs remain. We fine-tune \textbf{Qwen3.5-9B}~\citep{qwen359b} on these pairs for one epoch using the SFT objective in Eq.~\eqref{eq:sft_loss}. The resulting model is used as the Anchored PE for all subsequent caption construction and inference.

DiT fine-tuning uses candidate videos from \textbf{OpenVid-1M}~\citep{openvid}, \textbf{MiraData}~\citep{miradata}, \textbf{Video-UFO}~\citep{videoufo}, \textbf{MovieStory101}~\citep{moviestory101}, \textbf{LSMDC}~\citep{lsmdc}, and \textbf{Koala-36M}~\citep{koala36m}. Because the accompanying captions are often short and incomplete, using them directly risks providing lower-quality post-training supervision than the DiT encountered during pretraining. We therefore replace them with video-derived dense captions generated using \textbf{Qwen3.5-397B-A17B}, retain only pairs that pass video--caption consistency filtering, and apply further quality refinement, yielding 54K pairs. These filtered dense captions provide the shared, video-derived content sources processed by both caption operators. This video--caption set is disjoint from the data used to train the Anchored PE.
Each retained dense caption is processed by two operators. We implement the prompted rewriter using \textbf{Qwen3.5-397B-A17B} with the fixed six-part instruction, whereas the Anchored PE generates captions using the behavior learned in Step~1. Their outputs are used to fine-tune the Schema-Aligned and PE-Aligned DiTs, respectively. Both DiTs use the weighted flow-matching objective in Eq.~\eqref{eq:fm_loss}. The two fine-tuning settings share the videos, dense-caption sources, caption schema, preprocessing, optimization, and training budget; the only difference is which caption operator produces the DiT fine-tuning captions. Both DiTs are trained for one epoch using AdamW with an initial learning rate of $1\!\times\!10^{-5}$, weight decay of 0.01, and an effective batch size of 64, held constant for Wan2.2 and decayed by cosine to $1\!\times\!10^{-6}$ for LTX-2.3. For Wan2.2, fine-tuning updates the high-noise expert using 81-frame, 16-fps clips at 480p; for LTX-2.3, it updates the full DiT using 121-frame, 24-fps clips at 480p. Each run uses 32 NVIDIA H20 (141\,GB) GPUs across four nodes; further training details appear in the appendix.

\paragraph{Comparison protocol.}
Table~\ref{tab:main_results} assigns a distinct role to each of five settings. None is the user-prompt baseline. Official PE denotes the released prompt enhancer configuration of each generator, used unmodified: Qwen2.5-14B-Instruct for Wan2.2~\citep{wan22model} and Gemma 3-12B-IT for LTX-2.3~\citep{ltx23model}. Anchored PE with the Original DiT isolates Step~1; Anchored PE with the Schema-Aligned DiT is the matched control, whose DiT is adapted to the schema but not to the caption operator used at inference; and Anchored PE with the PE-Aligned DiT is CAPE-T2V with both steps applied.

For all metrics except VBench-2.0 Diversity, every PE-conditioned setting uses three rewrites sampled with distinct PE seeds. Crossing these rewrites with three DiT latent-noise seeds forms a $3{\times}3$ grid of nine videos per prompt. Each video is scored separately with the corresponding benchmark metric. On StoryEval, scoring one video uses three independent judge calls, each covering all of that prompt's events; the three calls are combined event-wise by unanimous voting into one binary decision per event. A prompt therefore uses $9{\times}3=27$ StoryEval judge calls. For each prompt, the resulting video-level metric scores or StoryEval per-event decisions are then averaged across the nine videos before the benchmark's official rule aggregates across prompts. For these metrics, the None setting has no PE rewrite and averages the three video-level results obtained with the same DiT latent-noise seeds. VBench-2.0 Diversity instead follows its separate official 20-video protocol: the None setting generates 20 videos from the original prompt, whereas each PE-conditioned setting generates 20 videos for each of its three rewrites, yielding 60 videos per original prompt. Within each benchmark and model family, all settings share the same DiT latent-noise seed sets and non-seed generation parameters; PE-conditioned settings additionally share the same PE sampling seed sets.

\subsection{Main Results}
The first three settings isolate inference-time prompt enhancement on the Original DiTs. Without DiT adaptation, the Anchored PE improves over the None user-prompt baseline in all six aggregate comparisons, showing that Step~1 already benefits both Original DiTs before adaptation. Pairing the Anchored PE with the PE-Aligned DiT further improves all six aggregate scores over using the same PE with the Original DiT, with gains from 0.92 points on T2V-CompBench to 4.6 points on StoryEval. DiT adaptation therefore adds to, rather than substitutes for, the gains from prompt enhancement.

More importantly, the final two settings hold the inference-time PE fixed and test whether using the deployed PE to construct DiT fine-tuning captions provides benefits beyond schema matching alone. CAPE-T2V outperforms the Schema-Aligned DiT across both model families and all three benchmarks. The gains on StoryEval, VBench-2.0, and T2V-CompBench are 1.6, 1.12, and 0.30 points for Wan2.2, and 1.4, 0.92, and 0.65 points for LTX-2.3, respectively. Since Wan2.2 and LTX-2.3 differ in architecture and fine-tuning scope, this consistent advantage indicates that the effect is not specific to one generator. On StoryEval, relative to each generator's official PE, the Anchored PE gains 17.1 points on Wan2.2 but loses 0.6 points on LTX-2.3, and the two official PEs differ sharply in their own contribution over direct user-prompt conditioning (+1.6 versus +24.6 points). The advantage over the Schema-Aligned DiT nevertheless holds for both, so it does not depend on how strong a generator's released PE is.

\subsection{PE--Caption Gap Analysis}
\label{sec:gap_analysis}
We instantiate the discrepancy $D$ from the Method by comparing the two DiT fine-tuning distributions, $P_{\mathrm{PE}}^{\mathrm{dense}}$ and $P_{\mathrm{rewriter}}^{\mathrm{dense}}$, with the inference-time reference distribution $P_{\mathrm{PE}}^{\mathrm{user}}$. To form the reference set, we apply the Anchored PE once with a fixed seed to every user prompt in StoryEval, VBench-2.0, and T2V-CompBench, producing one rewrite for each of 3,053 prompts.
A qualitative projection of these distributions is provided in
Figure~\ref{fig:supp_umap_visual_alignment} in the appendix.

\paragraph{Embedding discrepancy.}
Quantitative analysis instantiates $D$ as MMD$^2$ over 4096-dimensional L2-normalized representations from frozen Qwen3-VL-Embedding-8B~\citep{qwen3vlembedding}. We use the unbiased estimator~\citep{gretton2012kernel} with an RBF kernel; within each analysis, the bandwidth is selected by the median heuristic and held fixed across the two caption comparisons. The full-set comparison includes each 54K-caption DiT fine-tuning set and all 3,053 reference rewrites. A Qwen-token-length-matched control retains 3,000 texts per set, with matched token counts differing at most two. Table~\ref{tab:qwen_distribution_mmd} reports both comparisons.

\begin{table}[!t]
\centering
\footnotesize
\setlength{\tabcolsep}{4pt}
\begin{tabular*}{\columnwidth}{@{\extracolsep{\fill}}ccc@{}}
\toprule
\multirow{2}{*}{\textbf{DiT fine-tuning captions}} & \multicolumn{2}{c}{\textbf{MMD}$^2\downarrow$} \\
\cmidrule(lr){2-3}
& \textbf{Full set} & \textbf{Length matched} \\
\midrule
Prompted rewriter captions & 0.0758 & 0.0764 \\
Anchored PE captions & \textbf{0.0673} & \textbf{0.0682} \\
\bottomrule
\end{tabular*}
\caption{\textbf{Embedding discrepancy to rewrites of user prompts.} MMD$^2$ is computed between each set of DiT fine-tuning captions and the fixed-seed Anchored PE rewrites of all user prompts from StoryEval, VBench-2.0, and T2V-CompBench. Lower is better.}
\label{tab:qwen_distribution_mmd}
\end{table}

\begin{table}[!t]
\centering
\footnotesize
\setlength{\tabcolsep}{2pt}
\begin{tabular*}{\columnwidth}{@{\extracolsep{\fill}}cccc@{}}
\toprule
\textbf{DiT fine-tuning captions} & \textbf{ROUGE-L F1} & \textbf{BLEU-4} & \textbf{Qwen cosine} \\
\midrule
Prompted rewriter captions & 59.79 & 43.90 & 0.93467 \\
Anchored PE captions & 58.03 & 42.13 & 0.93610 \\
\bottomrule
\end{tabular*}
\caption{\textbf{Similarity to paired dense-caption sources over 54K examples.} Each DiT fine-tuning caption in the two sets is compared with its paired video-derived dense-caption source. ROUGE-L and BLEU-4 measure surface overlap on a 0--100 scale; Qwen cosine measures embedding similarity on a 0--1 scale.}
\label{tab:rewrite_source_proximity}
\end{table}

Anchored PE captions reduce MMD$^2$ by about 11\% relative to prompted rewriter captions in both the full-set and length-matched comparisons. The ordering remains unchanged after controlling for Qwen token length, indicating that length alone does not explain the result.

\subsection{Relation to Dense-Caption Sources}
To characterize how each caption operator transforms its shared video-derived source, we compare every derived caption with its paired dense caption over the full 54K set. Table~\ref{tab:rewrite_source_proximity} reports ROUGE-L F1~\citep{lin2004rouge} and corpus-level BLEU-4~\citep{papineni2002bleu} for lexical and sequence overlap, together with cosine similarity between frozen Qwen3-VL-Embedding-8B representations~\citep{qwen3vlembedding} for embedding-level semantic proximity.

Compared with prompted rewriter captions, Anchored PE captions have lower ROUGE-L and BLEU-4, indicating less direct reuse of the dense source's wording and local sequence structure. Their Qwen cosine similarity is nearly unchanged (0.93610 vs.\ 0.93467), indicating comparable embedding proximity to the paired dense-caption sources in the selected representation. Together with the lower MMD$^2$ to inference-time Anchored PE rewrites, this indicates that the measured gap reduction is not explained by greater lexical reuse of the shared dense-caption sources.

\section{Discussion}

\paragraph{Implications.}
The practical lesson is direct: when a prompt enhancer produces the text that conditions a generator at inference, reusing it to write the captions for generator adaptation can be more effective than relying on a separate caption operator that merely matches the schema. Schema matching remains useful---the Schema-Aligned DiT is a strong baseline---but captions written by the deployed PE retain a consistent advantage under this matched comparison. The two steps also contribute separately rather than substituting for each other, since anchoring the PE already improves both released generators and fine-tuning the DiT on that PE's captions adds further gains on every benchmark. Because the construction changes neither the generator architecture nor its training objective, it naturally extends to other generators equipped with inference-time PEs. Finally, the lower MMD$^2$ is not accompanied by greater lexical reuse of the dense sources: Anchored PE captions reuse less wording than prompted rewriter captions while remaining comparably close in embedding space. This pattern indicates that their smaller measured discrepancy to inference-time rewrites reflects the caption operator's realization rather than increased copying from the shared sources.

\paragraph{Evidence boundaries and limitations.}
Beyond each generator's released Official PE, we report no numbers for published prompt enhancers. Although the schema can be specified without recovering a generator's native captioning recipe, we evaluate a single instantiation, so behavior under other schemas remains open. We also do not ablate the three PE input constructions individually; our evidence therefore supports the resulting mixed-input PE but does not establish that each component is necessary. Our reading that the Anchored PE reorganizes its input rather than embellishing it rests on embedding and surface-overlap proximity to the dense-caption sources, not on fact-level verification against the videos. Finally, each DiT configuration is fine-tuned once, so we do not characterize variation across training runs.

\paragraph{Outlook.}
Table~\ref{tab:main_results} shows that the two generators' own PEs contribute very differently, which suggests a prediction worth testing more broadly: matching the DiT fine-tuning caption operator to the deployed PE should matter most where a generator's own PE is already strong, since that is where schema adaptation alone leaves the least headroom. A second question the construction raises is what makes a schema a good anchor---we choose one that exposes subjects, scene, camera, and style, but the anchor is a free parameter, and its content may determine how much of the gap can be closed. A third is whether the coupling can survive an updated enhancer: at present a new enhancer requires regenerating the fine-tuning captions and repeating adaptation, and cheaper incremental refresh would make the recipe practical for systems whose PEs change often.

\FloatBarrier

\section{Conclusion}

A DiT learns text--video correspondence from one distribution of conditioning text and is conditioned on another at inference, and sharing a caption schema does not remove the difference. CAPE-T2V couples the two distributions from outside the generator: the Anchored PE is trained on captioner-generated targets under the selected schema and then produces both the captions used for DiT fine-tuning and the rewritten prompts used at inference, without requiring access to a generator's native captioning recipe or joint PE--DiT training. Against a schema-aligned control whose fine-tuning captions are not produced by the deployed PE, this improves Wan2.2 and LTX-2.3 on all three benchmarks and narrows the measured gap, leaving each generator's architecture and objective unchanged.

% Finish all main-paper floats before the references.
\FloatBarrier
\bibliography{references}

@article{pos,
  title   = {{POS}: A Prompts Optimization Suite for Augmenting Text-to-Video Generation},
  author  = {Ma, Shijie and Xu, Huayi and Li, Mengjian and Geng, Weidong and Wang, Yaxiong and Wang, Meng},
  journal = {arXiv preprint arXiv:2311.00949},
  year    = {2023}
}

@inproceedings{instancecap,
  title     = {InstanceCap: Improving Text-to-Video Generation via Instance-aware Structured Caption},
  author    = {Fan, Tiehan and Nan, Kepan and Xie, Rui and Zhou, Penghao and Yang, Zhenheng and Fu, Chaoyou and Li, Xiang and Yang, Jian and Tai, Ying},
  booktitle = {Proceedings of the IEEE/CVF Conference on Computer Vision and Pattern Recognition},
  pages     = {28974--28983},
  year      = {2025}
}

@inproceedings{vpo,
  title     = {{VPO}: Aligning Text-to-Video Generation Models with Prompt Optimization},
  author    = {Cheng, Jiale and Lyu, Ruiliang and Gu, Xiaotao and Liu, Xiao and Xu, Jiazheng and Lu, Yida and Teng, Jiayan and Yang, Zhuoyi and Dong, Yuxiao and Tang, Jie and Wang, Hongning and Huang, Minlie},
  booktitle = {Proceedings of the IEEE/CVF International Conference on Computer Vision},
  pages     = {15636--15645},
  year      = {2025}
}

@inproceedings{promptavideo,
  title     = {Prompt-{A}-Video: Prompt Your Video Diffusion Model via Preference-Aligned {LLM}},
  author    = {Ji, Yatai and Zhang, Jiacheng and Wu, Jie and Zhang, Shilong and Chen, Shoufa and Ge, Chongjian and Sun, Peize and Chen, Weifeng and Shao, Wenqi and Xiao, Xuefeng and Huang, Weilin and Luo, Ping},
  booktitle = {Proceedings of the IEEE/CVF International Conference on Computer Vision},
  pages     = {18725--18735},
  year      = {2025}
}

@inproceedings{rapo,
  title     = {The Devil is in the Prompts: Retrieval-Augmented Prompt Optimization for Text-to-Video Generation},
  author    = {Gao, Bingjie and Gao, Xinyu and Wu, Xiaoxue and Zhou, Yujie and Qiao, Yu and Niu, Li and Chen, Xinyuan and Wang, Yaohui},
  booktitle = {Proceedings of the IEEE/CVF Conference on Computer Vision and Pattern Recognition},
  year      = {2025}
}

@article{rapopp,
  title   = {{RAPO++}: Cross-Stage Prompt Optimization for Text-to-Video Generation via Data Alignment and Test-Time Scaling},
  author  = {Gao, Bingjie and Ma, Qianli and Wu, Xiaoxue and Yang, Shuai and Lan, Guanzhou and Zhao, Haonan and Chen, Jiaxuan and Liu, Qingyang and Qiao, Yu and Chen, Xinyuan and Wang, Yaohui and Niu, Li},
  journal = {arXiv preprint arXiv:2510.20206},
  year    = {2025}
}

@article{avcdpo,
  title={{AVC-DPO}: Aligned Video Captioning via Direct Preference Optimization},
  author={Tang, Jiyang and Li, Hengyi and Du, Yifan and Zhao, Wayne Xin},
  journal={arXiv preprint arXiv:2507.01492},
  year={2025}
}

@inproceedings{videodpo,
  title     = {{VideoDPO}: Omni-Preference Alignment for Video Diffusion Generation},
  author    = {Liu, Runtao and Wu, Haoyu and Zheng, Ziqiang and Wei, Chen and He, Yingqing and Pi, Renjie and Chen, Qifeng},
  booktitle = {Proceedings of the IEEE/CVF Conference on Computer Vision and Pattern Recognition},
  pages     = {8009--8019},
  year      = {2025}
}

@inproceedings{promptist,
  title={Optimizing Prompts for Text-to-Image Generation},
  author={Hao, Yaru and Chi, Zewen and Dong, Li and Wei, Furu},
  booktitle={Advances in Neural Information Processing Systems (NeurIPS)},
  year={2023}
}

@article{promptrl,
  title={{PromptRL}: Prompt Matters in {RL} for Flow-Based Image Generation},
  author={Wang, Fu-Yun and Zhang, Han and Gharbi, Michael and Li, Hongsheng and Park, Taesung},
  journal={arXiv preprint arXiv:2602.01382},
  year={2026}
}

@inproceedings{promptcot,
  title     = {{PromptCoT}: Align Prompt Distribution via Adapted Chain-of-Thought},
  author    = {Yao, Junyi and Liu, Yijiang and Dong, Zhen and Guo, Mingfei and Hu, Helan and Keutzer, Kurt and Du, Li and Zhou, Daquan and Zhang, Shanghang},
  booktitle = {Proceedings of the IEEE/CVF Conference on Computer Vision and Pattern Recognition},
  pages     = {7027--7037},
  year      = {2024}
}

@article{dalle3,
  title={Improving Image Generation with Better Captions},
  author={Betker, James and Goh, Gabriel and Jing, Li and Brooks, Tim and Wang, Jianfeng and Li, Linjie and Ouyang, Long and Zhuang, Juntang and Lee, Joyce and Guo, Yufei and Manassra, Wesam and Dhariwal, Prafulla and Chu, Casey and Jiao, Yunxin and Ramesh, Aditya},
  journal={OpenAI Technical Report},
  year={2023}
}

@article{moviegen,
  title   = {Movie Gen: A Cast of Media Foundation Models},
  author  = {{Movie Gen Team}},
  journal = {arXiv preprint arXiv:2410.13720},
  year    = {2024}
}

@article{recap,
  title={A Picture is Worth a Thousand Words: Principled Recaptioning Improves Image Generation},
  author={Segalis, Eyal and Valevski, Dani and Lumen, Danny and Matias, Yossi and Leviathan, Yaniv},
  journal={arXiv preprint arXiv:2310.16656},
  year={2023}
}

@inproceedings{vc4vg,
  title={{VC4VG}: Optimizing Video Captions for Text-to-Video Generation},
  author={Du, Yang and Lin, Zhuoran and Song, Kaiqiang and Wang, Biao and Zheng, Zhicheng and Ge, Tiezheng and Zheng, Bo and Jin, Qin},
  booktitle={Conference on Empirical Methods in Natural Language Processing (EMNLP)},
  year={2025}
}

@inproceedings{miradata,
  title={{MiraData}: A Large-Scale Video Dataset with Long Durations and Structured Captions},
  author={Ju, Xuan and Gao, Yiming and Zhang, Zhaoyang and Yuan, Ziyang and Wang, Xintao and Zeng, Ailing and Xiong, Yu and Xu, Qiang and Shan, Ying},
  booktitle={Advances in Neural Information Processing Systems},
  volume={37},
  pages={48955--48970},
  year={2024}
}

@article{wan22,
  title={Wan: Open and Advanced Large-Scale Video Generative Models},
  author={{Wan Team}},
  journal={arXiv preprint arXiv:2503.20314},
  year={2025}
}

@article{hunyuanvideo15,
  title={{HunyuanVideo} 1.5 Technical Report},
  author={{Tencent Hunyuan Foundation Model Team}},
  journal={arXiv preprint arXiv:2511.18870},
  year={2025}
}

@misc{qwen35vl,
  title        = {{Qwen3.5-397B-A17B} Model Card},
  author       = {{Qwen Team}},
  howpublished = {Hugging Face model card},
  url          = {https://huggingface.co/Qwen/Qwen3.5-397B-A17B},
  year={2026},
  note         = {Accessed 2026-07-18}
}

@misc{qwen359b,
  title        = {{Qwen3.5-9B} Model Card},
  author       = {{Qwen Team}},
  howpublished = {Hugging Face model card},
  url          = {https://huggingface.co/Qwen/Qwen3.5-9B},
  year         = {2026},
  note         = {Accessed 2026-07-22}
}

@article{qwen3vlembedding,
  title={{Qwen3-VL-Embedding} and {Qwen3-VL-Reranker}: A Unified Framework for State-of-the-Art Multimodal Retrieval and Ranking},
  author={Li, Mingxin and Zhang, Yanzhao and Long, Dingkun and Chen, Keqin and Song, Sibo and Bai, Shuai and Yang, Zhibo and Xie, Pengjun and Yang, An and Liu, Dayiheng and Zhou, Jingren and Lin, Junyang},
  journal={arXiv preprint arXiv:2601.04720},
  year={2026}
}

@article{ltx2,
  title={{LTX-2}: Efficient Joint Audio-Visual Foundation Model},
  author={HaCohen, Yoav and Brazowski, Benny and Chiprut, Nisan and Bitterman, Yaki and Kvochko, Andrew and Berkowitz, Avishai and Shalem, Daniel and Lifschitz, Daphna and Moshe, Dudu and Porat, Eitan and Richardson, Eitan and Shiran, Guy and Chachy, Itay and Chetboun, Jonathan and Finkelson, Michael and Kupchick, Michael and Zabari, Nir and Guetta, Nitzan and Kotler, Noa and Bibi, Ofir and Gordon, Ori and Panet, Poriya and Benita, Roi and Armon, Shahar and Kulikov, Victor and Inger, Yaron and Shiftan, Yonatan and Melumian, Zeev and Farbman, Zeev},
  journal={arXiv preprint arXiv:2601.03233},
  year={2026}
}

@misc{ltx23model,
  title={{LTX-2.3} Model Card},
  author={{Lightricks}},
  howpublished={Hugging Face model card},
  url={https://huggingface.co/Lightricks/LTX-2.3},
  year={2026},
  note={Accessed 2026-07-18}
}

@misc{wan22model,
  title={{Wan2.2-T2V-A14B} Model Card},
  author={{Wan Team}},
  howpublished={Hugging Face model card},
  url={https://huggingface.co/Wan-AI/Wan2.2-T2V-A14B},
  year={2025},
  note={Accessed 2026-07-18}
}

@inproceedings{openvid,
  title={{OpenVid-1M}: A Large-Scale High-Quality Dataset for Text-to-Video Generation},
  author={Nan, Kepan and Xie, Rui and Zhou, Penghao and Fan, Tiehan and Yang, Zhenheng and Chen, Zhijie and Li, Xiang and Yang, Jian and Tai, Ying},
  booktitle={International Conference on Learning Representations},
  year={2025}
}

@inproceedings{videoufo,
  title={{VideoUFO}: A Million-Scale User-Focused Dataset for Text-to-Video Generation},
  author={Wang, Wenhao and Yang, Yi},
  booktitle={Advances in Neural Information Processing Systems},
  volume={38},
  year={2025}
}

@inproceedings{vidprom,
  title={{VidProM}: A Million-scale Real Prompt-Gallery Dataset for Text-to-Video Diffusion Models},
  author={Wang, Wenhao and Yang, Yi},
  booktitle={Advances in Neural Information Processing Systems},
  volume={37},
  year={2024}
}

@article{moviestory101,
  title={{StoryTeller}: Improving Long Video Description through Global Audio-Visual Character Identification},
  author={He, Yichen and Lin, Yuan and Wu, Jianchao and Zhang, Hanchong and Zhang, Yuchen and Le, Ruicheng},
  journal={arXiv preprint arXiv:2411.07076},
  year={2024}
}

@article{lsmdc,
  title={Movie Description},
  author={Rohrbach, Anna and Torabi, Atousa and Rohrbach, Marcus and Tandon, Niket and Pal, Christopher and Larochelle, Hugo and Courville, Aaron and Schiele, Bernt},
  journal={International Journal of Computer Vision},
  volume={123},
  number={1},
  pages={94--120},
  year={2017}
}

@inproceedings{koala36m,
  title={{Koala-36M}: A Large-scale Video Dataset Improving Consistency between Fine-grained Conditions and Video Content},
  author={Wang, Qiuheng and Shi, Yukai and Ou, Jiarong and Chen, Rui and Lin, Ke and Wang, Jiahao and Jiang, Boyuan and Yang, Haotian and Zheng, Mingwu and Tao, Xin and Yang, Fei and Wan, Pengfei and Zhang, Di},
  booktitle={Proceedings of the IEEE/CVF Conference on Computer Vision and Pattern Recognition},
  pages={8428--8437},
  year={2025}
}

@article{vbench2,
  title={{VBench-2.0}: Advancing Video Generation Benchmark Suite for Intrinsic Faithfulness},
  author={Zheng, Dian and Huang, Ziqi and Liu, Hongbo and Zou, Kai and He, Yinan and Zhang, Fan and Gu, Lulu and Zhang, Yuanhan and He, Jingwen and Zheng, Wei-Shi and Qiao, Yu and Liu, Ziwei},
  journal={arXiv preprint arXiv:2503.21755},
  year={2025}
}

@inproceedings{storyeval,
  title={Is Your World Simulator a Good Story Presenter? A Consecutive Events-Based Benchmark for Future Long Video Generation Models},
  author={Wang, Yiping and He, Xuehai and Wang, Kuan and Ma, Luyao and Yang, Jianwei and Wang, Shuohang and Du, Simon Shaolei and Shen, Yelong},
  booktitle={Proceedings of the IEEE/CVF Conference on Computer Vision and Pattern Recognition},
  pages={13629--13638},
  year={2025}
}

@inproceedings{t2vcompbench,
  title={{T2V-CompBench}: A Comprehensive Benchmark for Compositional Text-to-Video Generation},
  author={Sun, Kaiyue and Huang, Kaiyi and Liu, Xian and Wu, Yue and Xu, Zihan and Li, Zhenguo and Liu, Xihui},
  booktitle={Proceedings of the IEEE/CVF Conference on Computer Vision and Pattern Recognition},
  pages={8406--8416},
  year={2025}
}

@article{gretton2012kernel,
  title   = {A Kernel Two-Sample Test},
  author  = {Gretton, Arthur and Borgwardt, Karsten M. and Rasch, Malte J. and Sch{\"o}lkopf, Bernhard and Smola, Alexander},
  journal = {Journal of Machine Learning Research},
  volume  = {13},
  number  = {25},
  pages   = {723--773},
  year    = {2012}
}

@article{mcinnes2018umap,
  title   = {{UMAP}: Uniform Manifold Approximation and Projection for Dimension Reduction},
  author  = {McInnes, Leland and Healy, John and Melville, James},
  journal = {arXiv preprint arXiv:1802.03426},
  year    = {2018}
}

@inproceedings{papineni2002bleu,
  title={{BLEU}: A Method for Automatic Evaluation of Machine Translation},
  author={Papineni, Kishore and Roukos, Salim and Ward, Todd and Zhu, Wei-Jing},
  booktitle={Proceedings of the 40th Annual Meeting of the Association for Computational Linguistics},
  pages={311--318},
  year={2002}
}

@inproceedings{lin2004rouge,
  title={{ROUGE}: A Package for Automatic Evaluation of Summaries},
  author={Lin, Chin-Yew},
  booktitle={Text Summarization Branches Out},
  pages={74--81},
  year={2004}
}

@inproceedings{post2018sacrebleu,
  title={A Call for Clarity in Reporting {BLEU} Scores},
  author={Post, Matt},
  booktitle={Proceedings of the Third Conference on Machine Translation: Research Papers},
  pages={186--191},
  year={2018}
}

@inproceedings{koehn2004bootstrap,
  title={Statistical Significance Tests for Machine Translation Evaluation},
  author={Koehn, Philipp},
  booktitle={Proceedings of the 2004 Conference on Empirical Methods in Natural Language Processing},
  pages={388--395},
  year={2004}
}

% Include the appendix after the references in the same PDF.
\clearpage
% Appendix included by main.tex.
% This file is not a separate compilation entry point.

% Color encodes the caption role rather than the individual example: all Step~1
% SFT targets share one color, while the two Step~2 caption operators are distinct.
\definecolor{sfttargetcolor}{RGB}{42,153,125}
\definecolor{rewritercolor}{RGB}{190,112,94}
\definecolor{anchoredcolor}{RGB}{0,153,214}

% Representative text examples remain ordinary AAAI body text. The case title is
% typeset as its own ragged-right bold line so that long labels containing
% monospace clip identifiers do not stretch across a justified run-in heading.
\newenvironment{trainingcase}[1]
  {\par\vspace{6pt}\noindent{\bfseries\raggedright #1\par}%
   \nobreak\vspace{3pt}\noindent\ignorespaces}
  {\par\vspace{6pt}}

% Six-part captioner-generated targets are displayed with one numbered field per
% line in ragged-right italic, avoiding the stretched spacing that results when a
% long structured block is justified as a single paragraph.
\newenvironment{sixpartcaption}
  {\par\vspace{3pt}\begingroup\itshape\raggedright\setlength{\parindent}{0pt}%
   \setlength{\parskip}{2pt}}
  {\par\endgroup\vspace{3pt}}
\newcommand{\sixpartfield}[2]{\textbf{\itshape #1:}~#2\par}

% Lightweight flowable case blocks. Colored headings and rules retain the
% hierarchy while allowing TeX to paginate the body as ordinary text.
\newenvironment{captioncase}[2]{%
  \par\vspace{6pt}\begingroup
  \colorlet{caseframecolor}{#2}%
  \color{#2}\hrule height 0.65pt\vspace{4pt}%
  \noindent\textcolor{#2}{\bfseries #1:}\enspace\color{black}
}{%
  \par\vspace{4pt}\color{caseframecolor}\hrule height 0.65pt\endgroup\vspace{6pt}
}

\newenvironment{targetcase}[1]{%
  \par\vspace{6pt}\begingroup
  \colorlet{caseframecolor}{#1}%
  \color{#1}\hrule height 0.65pt\vspace{4pt}%
  \noindent\textcolor{#1}{\bfseries Complete captioner-generated target.}%
  \par\vspace{3pt}\color{black}
}{%
  \par\vspace{4pt}\color{caseframecolor}\hrule height 0.65pt\endgroup\vspace{6pt}
}

% Source inputs are separated visually from their long structured targets.
\newcommand{\caseinput}[1]{%
  \par\vspace{3pt}\noindent
  \begingroup\setlength{\fboxsep}{4pt}\setlength{\fboxrule}{0.4pt}%
  \fbox{\parbox{\dimexpr\linewidth-2\fboxsep-2\fboxrule\relax}{%
    \textbf{Input.}\enspace\textit{#1}}}%
  \endgroup\par\vspace{3pt}
}

% Behavioral notes use a compact plain frame.
\newcommand{\casebehavior}[1]{%
  \par\vspace{3pt}\noindent
  \begingroup\setlength{\fboxsep}{4pt}\setlength{\fboxrule}{0.35pt}%
  \fbox{\parbox{\dimexpr\linewidth-2\fboxsep-2\fboxrule\relax}{%
    \textbf{Target behavior.}\\[2pt]#1}}%
  \endgroup
  \par\vspace{3pt}
}

% Compact configuration tables use ordinary AAAI single-column floats so that
% surrounding text can continue when a table does not fit the current column.
% Wide result tables and qualitative figures remain two-column floats.
\newenvironment{configtable}
  {\begin{table}[t]\centering\small}
  {\end{table}}

% Qualitative composites contain four 480-pixel method strips followed by a
% 132-pixel prompt footer. Crop and label each method strip explicitly so that
% readers do not need to infer the row-to-method mapping.
\newcommand{\showcaselabel}[1]{%
  \begingroup\setlength{\fboxsep}{2.5pt}%
  \colorbox{black!58}{\parbox{\dimexpr\linewidth-2\fboxsep\relax}{%
    \color{white}\sffamily\bfseries\footnotesize #1}}%
  \endgroup\par\nobreak
}
\newcommand{\showcaserow}[2]{%
  \showcaselabel{#1}%
  \includegraphics[width=\linewidth]{#2}%
  \par\smallskip
}
\newcommand{\showcasepanel}[2]{%
  \begin{minipage}[t]{0.66\textwidth}
  \showcaserow{None + Original DiT}{#1.row1.jpg}%
  \showcaserow{Anchored PE + Original DiT}{#1.row2.jpg}%
  \showcaserow{Anchored PE + Schema-Aligned DiT}{#1.row3.jpg}%
  \showcaserow{CAPE-T2V (Anchored PE + PE-Aligned DiT)}{#1.row4.jpg}%
  {\centering\small #2\par}
  \end{minipage}%
}

\appendix
\section*{Appendix}
\addcontentsline{toc}{section}{Appendix}
\flushbottom

This appendix follows the terminology and notation of the main text and
provides additional results. In particular, \emph{prompted rewriter} denotes
the separately prompted schema-aligned rewriter used to construct captions for
the Schema-Aligned DiT; \emph{Anchored PE} denotes the frozen prompt enhancer
from Step~1; and \emph{CAPE-T2V} denotes the pairing of the Anchored PE with the
PE-Aligned DiT.

\section{Additional Implementation Details}

\paragraph{Step~1: Anchoring the PE to Captioner-Generated Targets.}
This step produces the Captioner-Anchored PE (Anchored PE), the frozen rewriter
used throughout our experiments. The final SFT pool contains 742,263 pairs:
486,497 pairs (65.54\%) use concise source caption or pseudo user prompt inputs,
and 255,766 pairs (34.46\%) use detailed source caption inputs. All pairs map to
six-part captioner-generated targets. The final mixture is constructed by
quota-based proportional sampling, with examples randomly sampled within each
bucket. Table~\ref{tab:supp_pe_training} summarizes the complete optimization
setup.

The three constructions begin from shared target candidates, but pair-level
availability and quality control are applied independently; the retained corpus
therefore contains three input types without enforcing balanced triplets.

\begin{configtable}
\begin{tabularx}{\linewidth}{@{}p{0.38\linewidth}X@{}}
\toprule
\textbf{Parameter} & \textbf{Setting} \\
\midrule
Training / validation pairs & 734,842 / 7,421 \\
Training input groups & 481,633 concise or pseudo user; 253,209 detailed \\
Model / framework & Qwen3.5-9B (9.410B parameters) / ms-swift Megatron \\
Trainable / frozen parameters & 8.954B language-model / approx. 456M visual-module parameters \\
Fine-tuning method & Full language-model tuning; no LoRA \\
Hardware / parallelism & 4 nodes with 8 NVIDIA H20 GPUs each; DP 32 and TP/PP/CP/EP 1 \\
Micro-batch / accumulation / global batch & 4 / 2 / 256 \\
Epochs / optimizer updates & 1 / 2,871 \\
Precision & BF16 computation; FP32 optimizer states and master parameters \\
Optimizer & Adam \\
Learning-rate schedule & Peak $1\!\times\!10^{-5}$; 5\% warmup (approx. 144 steps); cosine decay to 0 \\
Global / data-order seed & 42 / 42 \\
Maximum sequence length & 8192 tokens \\
Input handling & Lazy tokenization; no packing or length grouping; standard padded execution \\
Sampler end handling & Final global batch padded to 256 with 134 repeated instances \\
\bottomrule
\end{tabularx}
\caption{Step~1 Anchored PE training configuration. DP, TP, PP, CP, and EP denote data, tensor, pipeline, context, and expert parallelism, respectively.}
\label{tab:supp_pe_training}
\end{configtable}

\paragraph{Forward-pair construction and consistency filtering.}
The concise and detailed source caption constructions (Types~1 and 2) use a
shared pool constructed from MiraData, Video-UFO, and OpenVid. The
construction procedure validates the fixed six-part format---fields 1--6 in
the prescribed order, with one numbered field per line---removes placeholders
and duplicates, applies available upstream quality checks, and shuffles the
result.

All pairs then undergo a deterministic, text-only Qwen3.5-397B-A17B
non-contradiction check. The judge compares each input with its existing captioner-generated target without
observing video or regenerating text. It permits compatible elaboration but
rejects contradictions or core omissions in explicit content and camera
constraints. Failed or missing judge results are discarded.
From 500,000 initial pairs, 271,077 pass (54.22\%) and 228,923 are rejected
(45.78\%), leaving a 271,077-pair forward pool.

\paragraph{Reverse-pair construction and consistency filtering.}
The pseudo user prompt construction (Type~3) generates a pseudo user prompt from
a MiraData captioner-generated target. Pseudo user prompts are generated only
from targets whose corresponding detailed source captions have already passed
quality control. Prompts from VidProM serve only as few-shot examples of real
user language during generation and are not included in the training set used
for PE anchoring. The generated prompts then undergo a separate consistency
check, and generation failures or fallback queries are discarded. No additional
post-hoc intersection is applied. Of 330,313 generated examples, 257,432 pass
(77.94\%) and 72,881 are rejected (22.06\%). These forward and reverse counts
describe diagnostic construction pools and are not concatenated directly to form
the final SFT mixture.

\paragraph{Final SFT composition.}
The filtering counts above describe intermediate construction pools rather than two
files concatenated directly for training. Table~\ref{tab:supp_sft_composition}
reports the final mixture before the 1\% validation split. We apply
quota-based proportional sampling across the defined source--input buckets and
randomly sample examples within each bucket. The counts below are the resulting
post-sampling composition; the intermediate construction pools are not simply
concatenated.

\begin{configtable}
\begin{tabularx}{\linewidth}{@{}p{0.37\linewidth}X@{}}
\toprule
\textbf{Partition} & \textbf{Composition and count} \\
\midrule
Concise or pseudo user input & 263,927 filtered 500K-v3 query; 168,479 Video-UFO Story; 54,091 OpenVid VBench-style. Total: 486,497 (65.54\%). \\
Detailed source caption input & 131,808 MiraData; 62,150 OpenVid; 61,808 Video-UFO. Total: 255,766 (34.46\%). \\
\midrule
MiraData source total & 292,657 (39.43\%) across both input groups \\
Video-UFO source total & 279,419 (37.64\%) across both input groups \\
OpenVid source total & 170,187 (22.93\%) across both input groups \\
\midrule
All SFT pairs & 742,263 (100.00\%) before the validation split \\
\bottomrule
\end{tabularx}
\caption{Final Step~1 SFT composition after quota-based proportional sampling and before the 1\% validation split. Input-group and source-total rows are two complementary views of the same 742,263 pairs.}
\label{tab:supp_sft_composition}
\end{configtable}

The final arithmetic closes as
$263{,}927+168{,}479+54{,}091+131{,}808+62{,}150+61{,}808=742{,}263$ and
$734{,}842+7{,}421=742{,}263$. The intermediate filtering counts above are not
added to reconstruct the final mixture.

\paragraph{Qwen caption construction.}
Qwen3.5-397B-A17B produces the six-part captioner-generated targets used in Step~1 and the video-derived dense captions used in Step~2. A separate invocation of Qwen3.5-397B-A17B produces the prompted rewriter captions from the filtered dense captions using the fixed six-part instruction provided in the project repository (Section~\ref{sec:supp_prompt_availability}). The same model is used for the consistency and quality-control judgments in data construction. It runs in FP8 with vLLM and tensor parallelism 8 on one node with eight NVIDIA H20 GPUs.

\paragraph{Step~2: Fine-tuning the DiT on Anchored PE Captions.}
Wan2.2 and LTX-2.3 use the same final set of 54K video--dense-caption pairs.
Candidate videos are collected from OpenVid-1M, MiraData, Video-UFO,
MovieStory101, LSMDC, and Koala-36M; after filtering and quality-based
selection, its source composition is reported below. Table~\ref{tab:supp_dit_shared}
summarizes the settings shared by all matched DiT fine-tuning runs.

\begin{configtable}
\begin{tabularx}{\linewidth}{@{}p{0.39\linewidth}X@{}}
\toprule
\textbf{Parameter} & \textbf{Setting} \\
\midrule
Unique videos / padded instances & 54,000 / 54,016 \\
Epochs / data steps / optimizer updates & 1 / 1,688 / 844 \\
World size / hardware & 32 / 32 NVIDIA H20 GPUs \\
Micro-batch / accumulation / effective batch & 1 / 2 / 64 \\
Optimizer & AdamW, $\beta_1{=}0.9$, $\beta_2{=}0.999$, $\epsilon{=}10^{-8}$ \\
Initial learning rate / weight decay & $1\!\times\!10^{-5}$ / 0.01 \\
Precision / warmup / gradient clipping & BF16 / 0 steps / none \\
Memory and distributed optimization & Gradient checkpointing; DeepSpeed ZeRO-2 with optimizer CPU offload \\
Sampler end handling & \texttt{drop\_last=false}; 16 padding instances \\
Training seed / checkpoint interval & 42 / every 100 data steps \\
Training objective / auxiliary losses & Weighted flow-matching velocity objective / none \\
Matched-run difference & Caption operator only \\
\bottomrule
\end{tabularx}
\caption{Shared Step~2 DiT fine-tuning configuration. The Schema-Aligned and PE-Aligned runs within each model family use identical settings.}
\label{tab:supp_dit_shared}
\end{configtable}

Step~1 uses only text pairs. We exclude its source examples from the 54K Step~2
set by matching normalized video IDs and \texttt{file\_path} values, with hashes
covering renamed files. Benchmark IDs and prompts are likewise excluded from
training and model-selection data.

Each model uses the same video specification at training and inference. Within each model family, the
Schema-Aligned and PE-Aligned DiTs use identical videos, dense-caption sources,
preprocessing, optimization, and training budgets; only the final caption
operator differs.

\paragraph{Wan2.2-specific settings.}
Only the high-noise DiT expert is trainable; the text encoder, VAE, and
low-noise expert are frozen. Training uses timestep indices 0--416 (code range
$[0,417)$, normalized range 0.0--0.417) and a constant learning rate of
$1\!\times\!10^{-5}$. Videos are $832\!\times\!480$, 81 frames, and 16 fps.

\paragraph{LTX-2.3-specific settings.}
The Gemma text encoder, VAE, and audio components are frozen, and audio
training is disabled. Training spans all 1,000 timestep indices and uses cosine
decay from $1\!\times\!10^{-5}$ to $1\!\times\!10^{-6}$. For LTX-2.3,
fine-tuning updates the full DiT using 121-frame, 24-fps clips at 480p. Each run
uses 32 NVIDIA H20 (141\,GB) GPUs across four nodes.

\paragraph{DiT training objective.}
Both Wan2.2 and LTX-2.3 use the weighted flow-matching velocity objective
defined in the main paper rather than an $\epsilon$-prediction objective.
Forward computation uses BF16, while predictions and targets are cast to FP32
for loss computation. Scheduler weights are normalized to have mean one. No
perceptual, CLIP, caption-generation, reconstruction, or other auxiliary loss
is used. Although the implementation names the prediction variable
\texttt{noise\_pred}, its target is the velocity $v^*=\epsilon-z_0$.

\paragraph{Video preprocessing.}
We retain only source videos that are at least 5 s but shorter than 10 s and
whose native resolution exceeds 480p. Retained videos are resized to
$832\!\times\!480$ (width$\times$height); training logs record this as
$480\!\times\!832$ in height$\times$width order. For clips longer than the
target duration, we use temporal downsampling rather than random temporal
cropping, looping, or frame padding. A second resampling pass creates the
model-specific MP4 files described above, so
both DiT families retain the same source-video identity while using their
respective native temporal specifications.

\paragraph{Dense-caption consistency filtering.}
After dense-caption generation, a Qwen3.5-397B-A17B video--caption faithfulness judge evaluates
every substantive caption claim against the source video and returns a Boolean
\texttt{consistent} decision; 65.98\% pass. Visual-quality filtering retains
87,609 of 94,887 inputs and rejects 7,278 still, zoom-only, first-frame-text,
or overlapping cases, yielding 87,609 caption-complete examples. Selection
requires quality at least 30 (20,779 fall below it) and captions of at least
eight tokens. Additional deduplication and distribution-balancing filters are
then applied before final selection.
Only selected captions are passed to the prompted rewriter and Anchored PE.
The final 54,000 unique videos comprise 18,383 OpenVid-1M, 1,370
MovieStory101, 2,097 LSMDC, and 32,150 Koala-36M examples. The 0.48
source-share target is a soft preference rather than a hard cap, so Koala-36M
may contribute $32{,}150/54{,}000=59.54\%$. The complete consistency-judge
prompt is provided in the project repository
(Section~\ref{sec:supp_prompt_availability}).
MiraData and Video-UFO contribute to the candidate pool, but no examples from
either source remain after the combined disjointness, quality, deduplication,
and balancing selection; the final 54K set therefore contains the four sources
listed above.

\section{Benchmark Inference Protocol}

\paragraph{Prompt-enhancer inference.}
StoryEval, VBench-2.0, and T2V-CompBench use the same Anchored PE and six-part system prompt at inference. StoryEval sends each official prompt directly to the Anchored PE; no additional dense-caption model is applied first. For each input, the Anchored PE produces three complete six-part outputs, rather than only the Dense Caption field, to condition the video generator. For the text-distribution diagnostics, we apply the same Anchored PE and six-part template once with sampling seed 42 to every user prompt in the complete StoryEval, VBench-2.0, and T2V-CompBench evaluation sets and pool the resulting inference-time PE outputs.

For StoryEval's 423 prompts, decoding uses temperature 0.7, top-$p$ 0.8,
top-$k$ 20, thinking disabled, and a 1,024-token limit. We use concurrency 64,
a 120-second request timeout, and store the complete six-part
\texttt{prompt} output.

The corresponding preprocessing implementation is included in the project
repository.

\paragraph{Video generation.}
Generation settings are held fixed across compared settings within each
benchmark and model family. Wan2.2 generates $832\!\times\!480$ videos with 81
frames at 16 fps using 40 FlowUniPC multistep iterations, flow shift 12.0, CFG
3.0 for the low-noise expert and 4.0 for the high-noise expert, expert boundary
0.875, and BF16 precision. LTX-2.3 generates $832\!\times\!480$ videos with 121
frames at 24 fps using 30 DiffSynth FlowMatchScheduler steps, CFG 4.0, and BF16
precision. Its spatial tiles are 512 pixels with 128-pixel overlap; temporal
tiles are 128 frames with 24-frame overlap.

Each DiT configuration is trained once with training seed 42. At evaluation,
PE sampling seeds 42, 666, and 888 are crossed with DiT latent-noise seeds 42,
666, and 888 to produce the $3{\times}3$ grid; None uses the three latent-noise seeds.
Scores average all nine PE-conditioned videos or three None videos. For an
Official PE, only the stochastic seed changes; its released model, template,
decoding, and other settings remain unmodified. VBench-2.0 Diversity instead
uses DiT seeds 0--19 for each rewrite and averages the three official scores;
None uses one 20-video set. Compared settings share seeds within each benchmark
and model family.

\paragraph{StoryEval judge configuration.}
Each generated video receives three independent GPT judge calls under
StoryEval's official strict mode. An event receives credit for a video only if
all three judgments mark it as completed. This gives $9\times3=27$ judge calls
per prompt for each PE-conditioned setting and $3\times3=9$ calls for None. We
average the resulting video-level decisions and then apply StoryEval's official
aggregation to obtain category-specific and overall completion rates. Thus, the
three judge calls determine one decision for a video; they are not additional
video samples and are not averaged as such. The judge backend and execution
controls are runtime configuration rather than constants of the evaluation
protocol. For the reported experiments, GPT-5.5 was supplied as the judge model
at runtime. In the released evaluation entry point, the judge model is provided
through a command-line argument or environment configuration; the defaults are
\texttt{repeat-time=3}, \texttt{num-workers=32}, and
\texttt{openai-retry=4}. Worker count, retry behavior, service endpoint, and
credentials may therefore be adjusted to the execution environment without
changing the three-call unanimity rule or the reported score aggregation. The
strict-mode aggregation uses \texttt{vote\_type=1}, \texttt{null\_type=0},
\texttt{if\_reeval=0}, and \texttt{if\_reeval\_null=1}.

The Overall column in the main paper reports StoryEval's official completion
rate aggregated over all evaluation prompts; it is not computed as an
unweighted mean of the displayed category and difficulty columns. The
evaluation implementation is included in the project repository;
service endpoints, credentials, user identifiers, and other
infrastructure-specific authentication fields are intentionally omitted.

\section{Complete VBench-2.0 Results}
Table~\ref{tab:supp_vbench2_dimensions} reports the five VBench-2.0 dimension
scores and their official average for Wan2.2 and LTX-2.3.

\paragraph{Compared settings.}
None directly conditions the Original DiT. Official PE uses each generator's
released enhancer with the Original DiT. Anchored PE isolates Step~1 on the
Original DiT. The final two settings both use the Anchored PE at inference;
their DiTs differ only in whether the fine-tuning captions come from the
prompted rewriter or the Anchored PE.

\begin{table*}[t]
\centering
\small
\begin{tabularx}{0.98\textwidth}{@{}p{0.18\textwidth}*{5}{>{\centering\arraybackslash}X}@{}}
\toprule
\textbf{Dimension} &
\shortstack{\textbf{None}\\\textbf{Original DiT}} &
\shortstack{\textbf{Official PE}\\\textbf{Original DiT}} &
\shortstack{\textbf{Anchored PE}\\\textbf{Original DiT}} &
\shortstack{\textbf{Anchored PE}\\\textbf{Schema-Aligned DiT}} &
\shortstack{\textbf{Anchored PE}\\\textbf{PE-Aligned DiT}\\\textbf{(CAPE-T2V)}} \\
\midrule
\multicolumn{6}{l}{\textbf{Wan2.2}} \\
Creativity & \underline{51.72\%} & 45.20\% & 49.95\% & 51.26\% & \textbf{53.19\%} \\
Commonsense & 66.32\% & 62.86\% & \underline{72.35\%} & 72.31\% & \textbf{74.95\%} \\
Controllability & 29.80\% & 31.53\% & \textbf{48.12\%} & 46.91\% & \underline{47.52\%} \\
Human fidelity & 79.45\% & \textbf{83.64\%} & 79.64\% & \underline{79.95\%} & 79.91\% \\
Physics & 47.76\% & 55.66\% & 54.28\% & \underline{58.20\%} & \textbf{58.69\%} \\
\midrule
Average & 55.01\% & 55.78\% & 60.87\% & \underline{61.73\%} & \textbf{62.85\%} \\
\midrule
\multicolumn{6}{l}{\textbf{LTX-2.3}} \\
Creativity & 43.41\% & \underline{46.70\%} & 45.98\% & 45.85\% & \textbf{48.10\%} \\
Commonsense & 61.09\% & \textbf{75.83\%} & 73.20\% & 73.51\% & \underline{75.11\%} \\
Controllability & 25.60\% & 37.67\% & \textbf{44.97\%} & 43.13\% & \underline{43.82\%} \\
Human fidelity & \textbf{91.87\%} & 84.07\% & \underline{84.58\%} & 83.34\% & 83.09\% \\
Physics & 42.42\% & \textbf{59.45\%} & 53.77\% & 57.32\% & \underline{57.63\%} \\
\midrule
Average & 52.88\% & \underline{60.74\%} & 60.50\% & 60.63\% & \textbf{61.55\%} \\
\bottomrule
\end{tabularx}
\normalsize
\caption{VBench-2.0 dimension scores (\%). PE-conditioned entries use the $3{\times}3$ PE--DiT seed grid; Diversity follows the separate three-by-20-video protocol. Higher is better; bold and underline denote the best and second-best results within each model family.}
\label{tab:supp_vbench2_dimensions}
\end{table*}

\paragraph{Interpretation.}
CAPE-T2V gives the highest VBench-2.0 average for both generators. The
stepwise comparison first shows the benefit of Step~1: Anchored PE raises the
Original-DiT average from 55.01\% to 60.87\% on Wan2.2 and from 52.88\% to
60.50\% on LTX-2.3. Applying Step~2 then raises these averages to 62.85\% and
61.55\%, respectively. The matched comparison isolates the caption operator:
both final settings use the Anchored PE at inference, while CAPE-T2V raises
the Schema-Aligned DiT average from 61.73\% to 62.85\% on Wan2.2 and from
60.63\% to 61.55\% on LTX-2.3. This matched gain is broad rather than driven by a single category: CAPE-T2V
improves Creativity, Commonsense, Controllability, and Physics on both model
families, while Human Fidelity changes only marginally (79.95\% to 79.91\%
and 83.34\% to 83.09\%). This pattern supports the main-paper claim that the
two-sided alignment provides an additional benefit beyond schema matching alone.

\section{Complete T2V-CompBench Results}
Table~\ref{tab:supp_t2vcomp_dimensions} reports all seven T2V-CompBench
dimensions and their averages for Wan2.2 and LTX-2.3. Each dimension contains
200 prompts, giving 1,400 evaluation prompts per model family.

\paragraph{Compared settings.}
The five columns follow the main paper: None uses the Original DiT without
rewriting; Official PE is the released-system reference; Anchored PE isolates
Step~1; Schema-Aligned DiT and CAPE-T2V both use the Anchored PE at inference
and differ only in the caption operator used for DiT fine-tuning.

\begin{table*}[t]
\centering
\small
\begin{tabularx}{0.98\textwidth}{@{}p{0.18\textwidth}*{5}{>{\centering\arraybackslash}X}@{}}
\toprule
\textbf{Dimension} &
\shortstack{\textbf{None}\\\textbf{Original DiT}} &
\shortstack{\textbf{Official PE}\\\textbf{Original DiT}} &
\shortstack{\textbf{Anchored PE}\\\textbf{Original DiT}} &
\shortstack{\textbf{Anchored PE}\\\textbf{Schema-Aligned DiT}} &
\shortstack{\textbf{Anchored PE}\\\textbf{PE-Aligned DiT}\\\textbf{(CAPE-T2V)}} \\
\midrule
\multicolumn{6}{l}{\textbf{Wan2.2}} \\
Consistent attribute & 84.05\% & 89.75\% & \underline{91.79\%} & \textbf{91.80\%} & 91.04\% \\
Dynamic attribute & 13.26\% & 8.97\% & 24.41\% & \underline{28.63\%} & \textbf{30.62\%} \\
Spatial relationship & 61.63\% & 63.40\% & \underline{70.25\%} & 68.63\% & \textbf{70.43\%} \\
Motion binding & 30.40\% & 31.80\% & 43.03\% & \textbf{49.06\%} & \underline{48.82\%} \\
Action binding & 77.07\% & \underline{81.26\%} & \textbf{86.13\%} & 80.44\% & 81.02\% \\
Object interaction & 70.52\% & 79.94\% & \textbf{85.54\%} & \underline{82.35\%} & 81.85\% \\
Numeracy & 54.38\% & 60.23\% & 59.05\% & \textbf{63.62\%} & \underline{62.83\%} \\
\midrule
Mean & 55.90\% & 59.34\% & 65.74\% & \underline{66.36\%} & \textbf{66.66\%} \\
\midrule
\multicolumn{6}{l}{\textbf{LTX-2.3}} \\
Consistent attribute & 79.05\% & 86.30\% & 87.02\% & \underline{88.48\%} & \textbf{88.50\%} \\
Dynamic attribute & 10.85\% & 20.78\% & 24.83\% & \underline{26.88\%} & \textbf{28.97\%} \\
Spatial relationship & 53.02\% & 61.68\% & \underline{70.62\%} & 68.81\% & \textbf{70.65\%} \\
Motion binding & 27.00\% & \textbf{34.91\%} & 33.36\% & 34.17\% & \underline{34.68\%} \\
Action binding & 52.04\% & \textbf{76.13\%} & 69.50\% & 71.75\% & \underline{71.85\%} \\
Object interaction & 46.93\% & 72.70\% & \textbf{76.09\%} & \underline{76.05\%} & 75.95\% \\
Numeracy & 28.11\% & 47.39\% & 46.52\% & \underline{53.17\%} & \textbf{53.26\%} \\
\midrule
Mean & 42.43\% & 57.13\% & 58.28\% & \underline{59.90\%} & \textbf{60.55\%} \\
\bottomrule
\end{tabularx}
\normalsize
\caption{T2V-CompBench dimension scores (\%). Model families and settings follow the organization of the main-results table. PE-conditioned entries use the $3{\times}3$ PE--DiT seed grid; None uses three DiT latent-noise seeds. Higher is better; bold and underline denote the best and second-best results within each model family.}
\label{tab:supp_t2vcomp_dimensions}
\end{table*}

\paragraph{Interpretation.}
CAPE-T2V gives the highest T2V-CompBench mean for both generators. Step~1
raises the Original-DiT mean from 55.90\% to 65.74\% on Wan2.2 and from
42.43\% to 58.28\% on LTX-2.3. Applying Step~2 further raises the means to
66.66\% and 60.55\%, respectively. In the matched comparison, CAPE-T2V
improves over the Schema-Aligned DiT from 66.36\% to 66.66\% on Wan2.2 and
from 59.90\% to 60.55\% on LTX-2.3. The dimension-level changes are not
uniform: both model families improve on dynamic attributes, spatial
relationships, and action binding, with the largest shared gains appearing in
dynamic attributes and spatial relationships. Wan2.2 shows small decreases on
several other dimensions, whereas LTX-2.3 improves on six of the seven
dimensions. The positive matched averages therefore support the benefit of
two-sided conditioning alignment without implying universal per-dimension
dominance.

% Defer the qualitative showcase to the end of the appendix. Keeping these
% six full-width floats out of the middle prevents their two-column float queue
% from affecting the analytical sections that follow.
\newcommand{\suppshowcases}{%
\section{Additional Qualitative Showcase}
Figures~\ref{fig:supp_showcase_ordered_ab} and~\ref{fig:supp_showcase_ordered_cd}
show ordered execution; Figures~\ref{fig:supp_showcase_dynamics_ef}
and~\ref{fig:supp_showcase_dynamics_gh} show dynamic control; and
Figures~\ref{fig:supp_showcase_interaction_ij}
and~\ref{fig:supp_showcase_interaction_kl} show subject--object interaction.
Together, these figures contain 12 manually reviewed cases. Each case presents four separately cropped and
labeled $1{\times}4$ keyframe strips: None with the Original DiT, Anchored PE
with the Original DiT, Anchored PE with the Schema-Aligned DiT, and CAPE-T2V.
The matched comparison is the last two strips, which share the Anchored PE and
schema but differ in the DiT fine-tuning caption operator. The exact original
benchmark prompt is printed below each case. Keyframes are selected
to expose events or failure modes and are not used for benchmark scoring.

\begin{figure*}[!t]
\centering
\showcasepanel{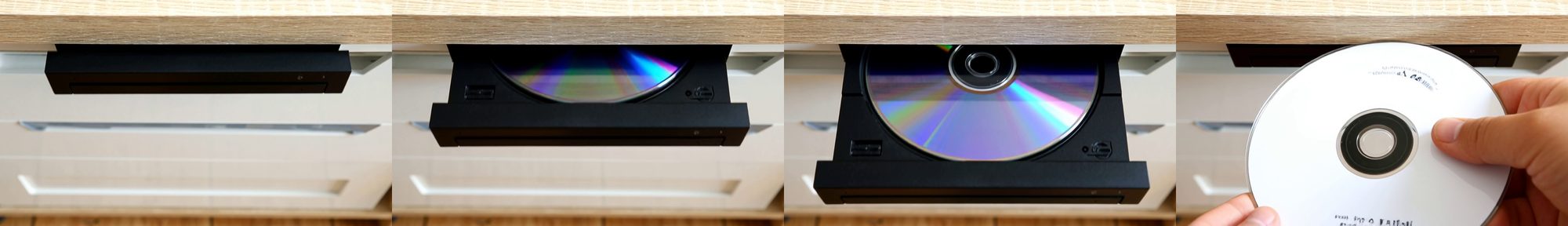}%
  {(a) StoryEval \#2. \textbf{Original prompt:} ``A CD tray opens, a disc is placed inside, and then the tray closes.''}\par\vspace{2mm}
\showcasepanel{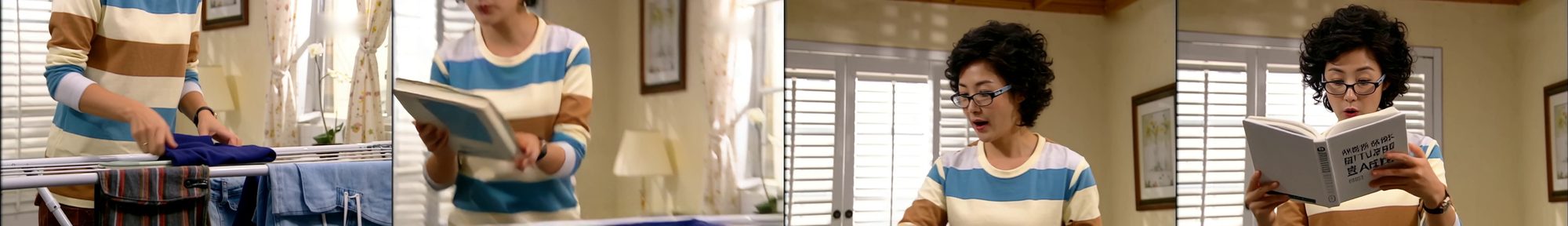}%
  {(b) VBench-2.0 \#628. \textbf{Original prompt:} ``A person is folding clothes, then they suddenly get up and start reading a book.''}
\caption{\textbf{Ordered multi-event execution (I).} Each model strip is
cropped from the corresponding row of the original composite and labeled
explicitly. In (a), None reverses the insertion sequence, Anchored PE
omits the initially closed tray state, and Schema-Aligned DiT introduces a disc
without stable placement. CAPE-T2V most clearly renders closed--open--insertion--closed.
In (b), None and Schema-Aligned DiT change the person, while Anchored PE is more
abrupt; CAPE-T2V better covers folding, rising, opening the book, and reading.}
\label{fig:supp_showcase_ordered_ab}
\end{figure*}

\begin{figure*}[!t]
\centering
\showcasepanel{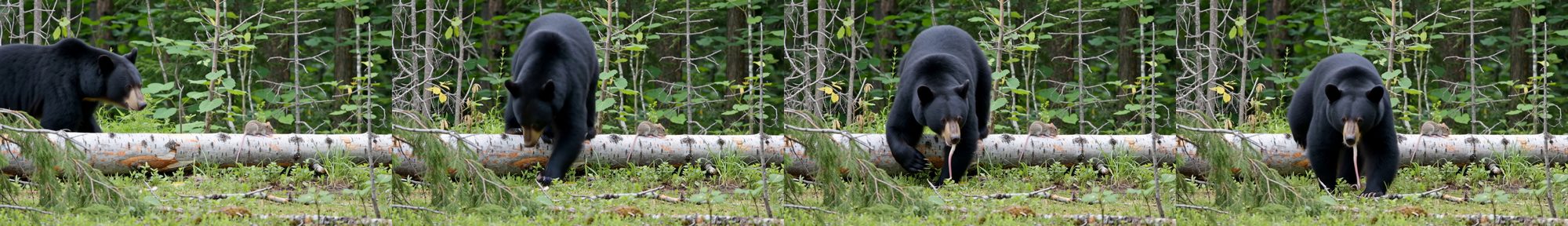}%
  {(c) StoryEval \#13. \textbf{Original prompt:} ``A bear pushes down a tree stump, finds food, catches a mouse, and eats it.''}\par\vspace{2mm}
\showcasepanel{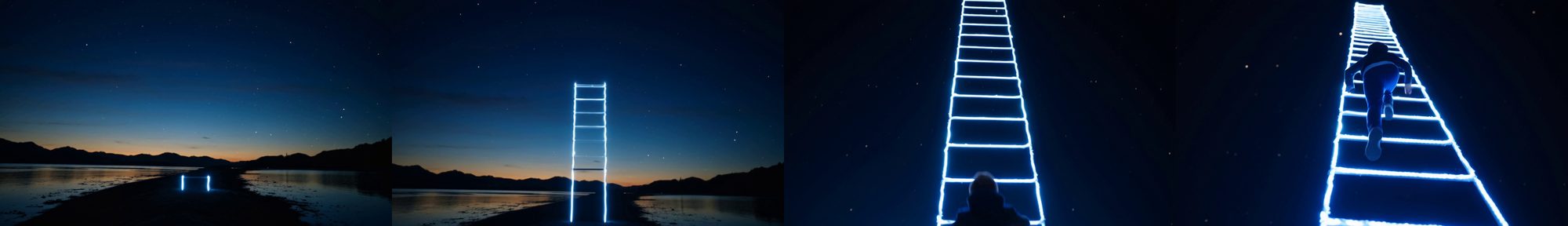}%
  {(d) StoryEval \#255. \textbf{Original prompt:} ``A moonbeam shines down, solidifies into a shining ladder, and then someone climbs up it.''}
\caption{\textbf{Ordered multi-event execution (II).} In (c), None omits the
mouse, Anchored PE stops after the reveal, and Schema-Aligned DiT is less stable
around capture and consumption; CAPE-T2V provides the broadest coverage, though
the final capture and eating remain visually ambiguous. In (d), None gives an
abrupt ladder/climber configuration, Anchored PE is nearly static, and
Schema-Aligned DiT lacks sustained climbing; CAPE-T2V best preserves the
moonbeam--ladder--upward-motion progression.}
\label{fig:supp_showcase_ordered_cd}
\end{figure*}

\begin{figure*}[!t]
\centering
\showcasepanel{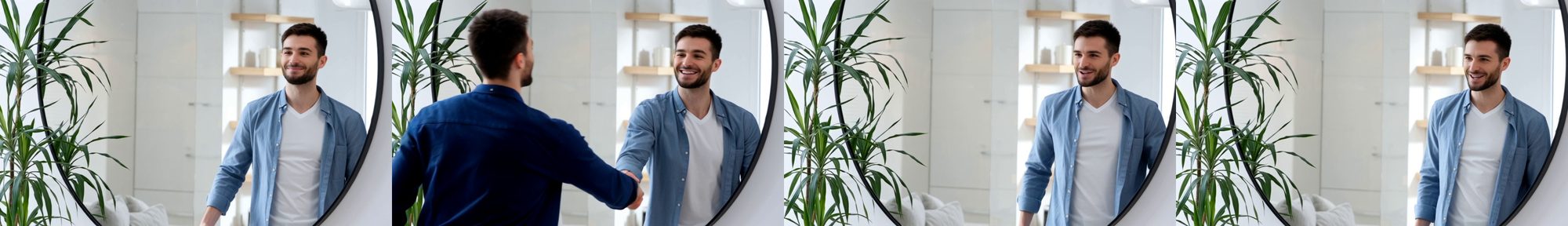}%
  {(e) StoryEval \#232. \textbf{Original prompt:} ``A man shakes hands with his reflection in the mirror, and then the reflection walks away.''}\par\vspace{2mm}
\showcasepanel{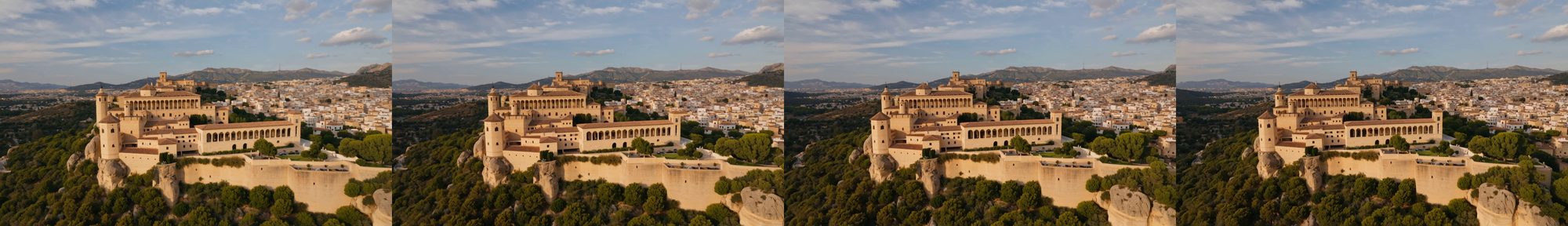}%
  {(f) VBench-2.0 \#1. \textbf{Original prompt:} ``Alhambra, First-person perspective, oblique shot, airborne dolly movement.''}
\caption{\textbf{Dynamic state and trajectory control (I).} In (e), None keeps
the reflection static, while Anchored PE and Schema-Aligned DiT let the second
figure enter the room; CAPE-T2V keeps the action in the mirror and ends with an
empty reflection. In (f), None is nearly static, Anchored PE rolls and pitches,
and Schema-Aligned DiT circles with occlusion; CAPE-T2V shows the clearest
forward path through sustained parallax.}
\label{fig:supp_showcase_dynamics_ef}
\end{figure*}

\begin{figure*}[!t]
\centering
\showcasepanel{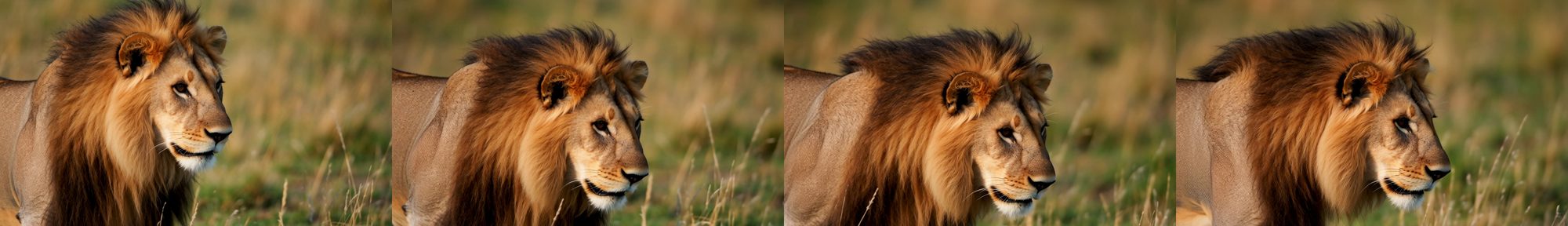}%
  {(g) VBench-2.0 \#281. \textbf{Original prompt:} ``A lion changes from big to small.''}\par\vspace{2mm}
\showcasepanel{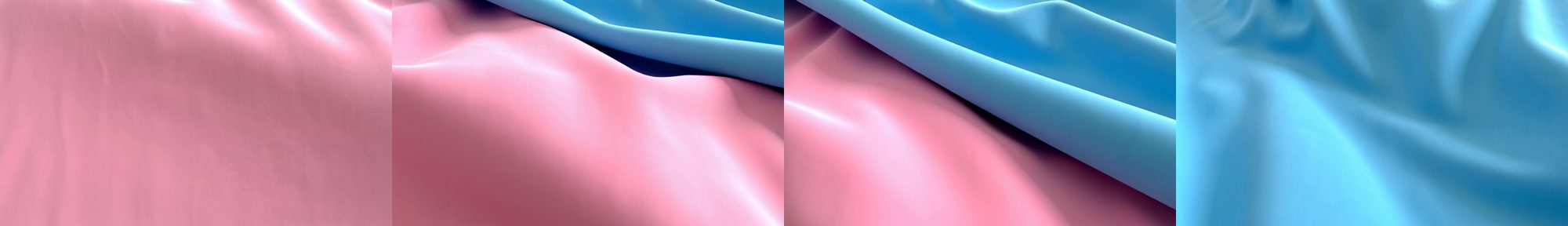}%
  {(h) VBench-2.0 \#290. \textbf{Original prompt:} ``A piece of fabric changes from pink to blue.''}
\caption{\textbf{Dynamic state and trajectory control (II).} In (g), None shows
little scale change, Anchored PE confounds scale with posture, and
Schema-Aligned DiT weakens body consistency; CAPE-T2V gives the smoothest
large-to-small trajectory while retaining identity. In (h), None uses unstable
close framing and the middle settings mix competing colors, whereas CAPE-T2V
keeps one bounded fabric panel through pink, intermediate purple, and blue
states.}
\label{fig:supp_showcase_dynamics_gh}
\end{figure*}

\begin{figure*}[!t]
\centering
\showcasepanel{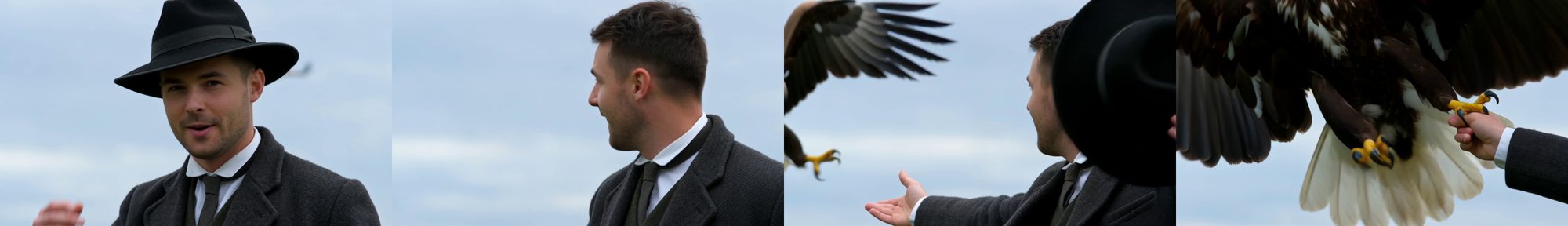}%
  {(i) StoryEval \#236. \textbf{Original prompt:} ``A man takes off his hat, throws it into the air, and then it is taken by a passing eagle.''}\par\vspace{2mm}
\showcasepanel{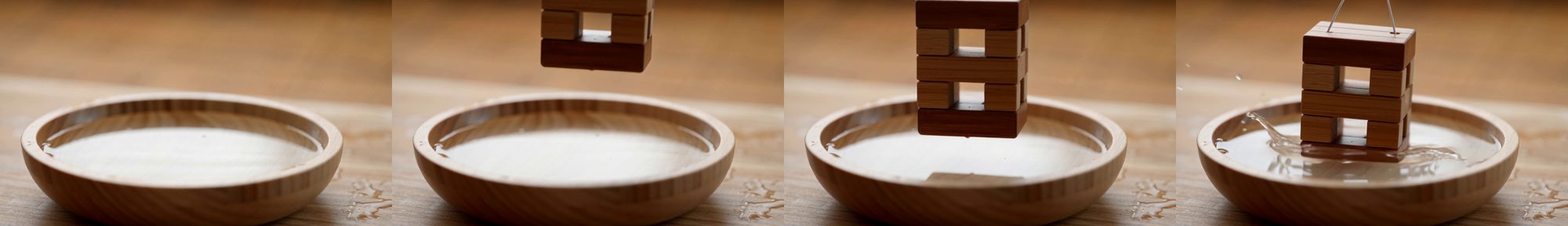}%
  {(j) VBench-2.0 \#915. \textbf{Original prompt:} ``A wooden toy is placed gently on the surface of a small bowl of water.''}
\caption{\textbf{Interaction and subject--object binding (I).} In (i), None
obscures the hat transfer, Anchored PE separates the eagle and hat, and
Schema-Aligned DiT lacks a stable carried endpoint; CAPE-T2V makes the removal,
throw, approach, and final eagle--hat binding most visible. In (j), None keeps
the structure attached, Anchored PE does not complete release, and
Schema-Aligned DiT keeps a hand-like contact; CAPE-T2V best shows lowering,
contact, release, and independent floating.}
\label{fig:supp_showcase_interaction_ij}
\end{figure*}

\begin{figure*}[!t]
\centering
\showcasepanel{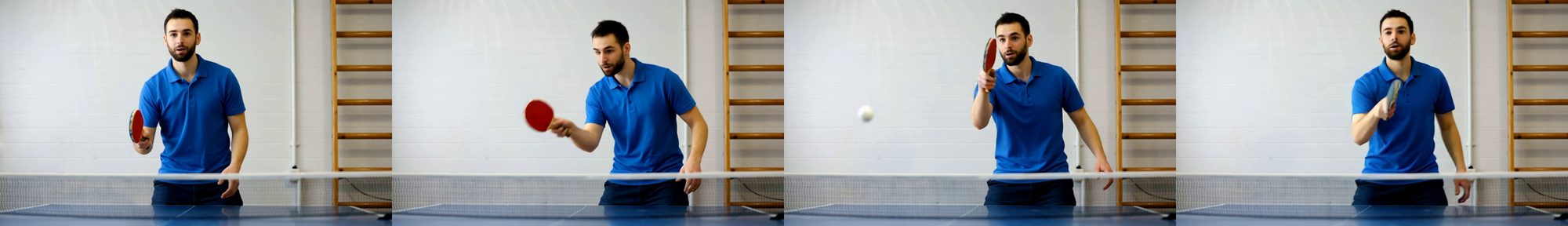}%
  {(k) VBench-2.0 \#421. \textbf{Original prompt:} ``A man is playing ping-pong.''}\par\vspace{2mm}
\showcasepanel{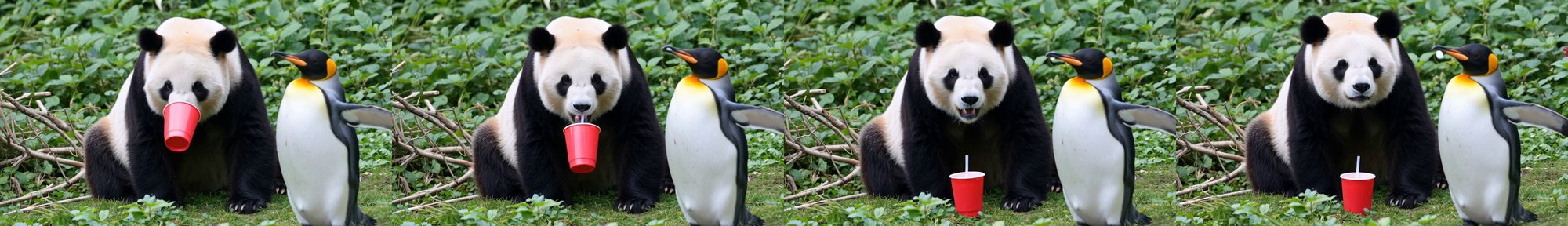}%
  {(l) T2V-CompBench \#189. \textbf{Original prompt:} ``A panda drinks tea while a penguin eats cookies.''}
\caption{\textbf{Interaction and subject--object binding (II).} In (k), all
methods produce a recognizable ping-pong scene, so the useful comparison is
fine-grained continuity: Anchored PE and Schema-Aligned DiT more often lose the
ball or hand--paddle relation, while CAPE-T2V keeps the interaction more
readable. In (l), CAPE-T2V most clearly attaches drinking to the panda and
eating to the penguin, supporting the subject--action binding claim.}
\label{fig:supp_showcase_interaction_kl}
\end{figure*}
}

\FloatBarrier
\flushbottom
\section{Text-Distribution Analysis}

\begin{figure}[!t]
\centering
\includegraphics[width=0.96\columnwidth]{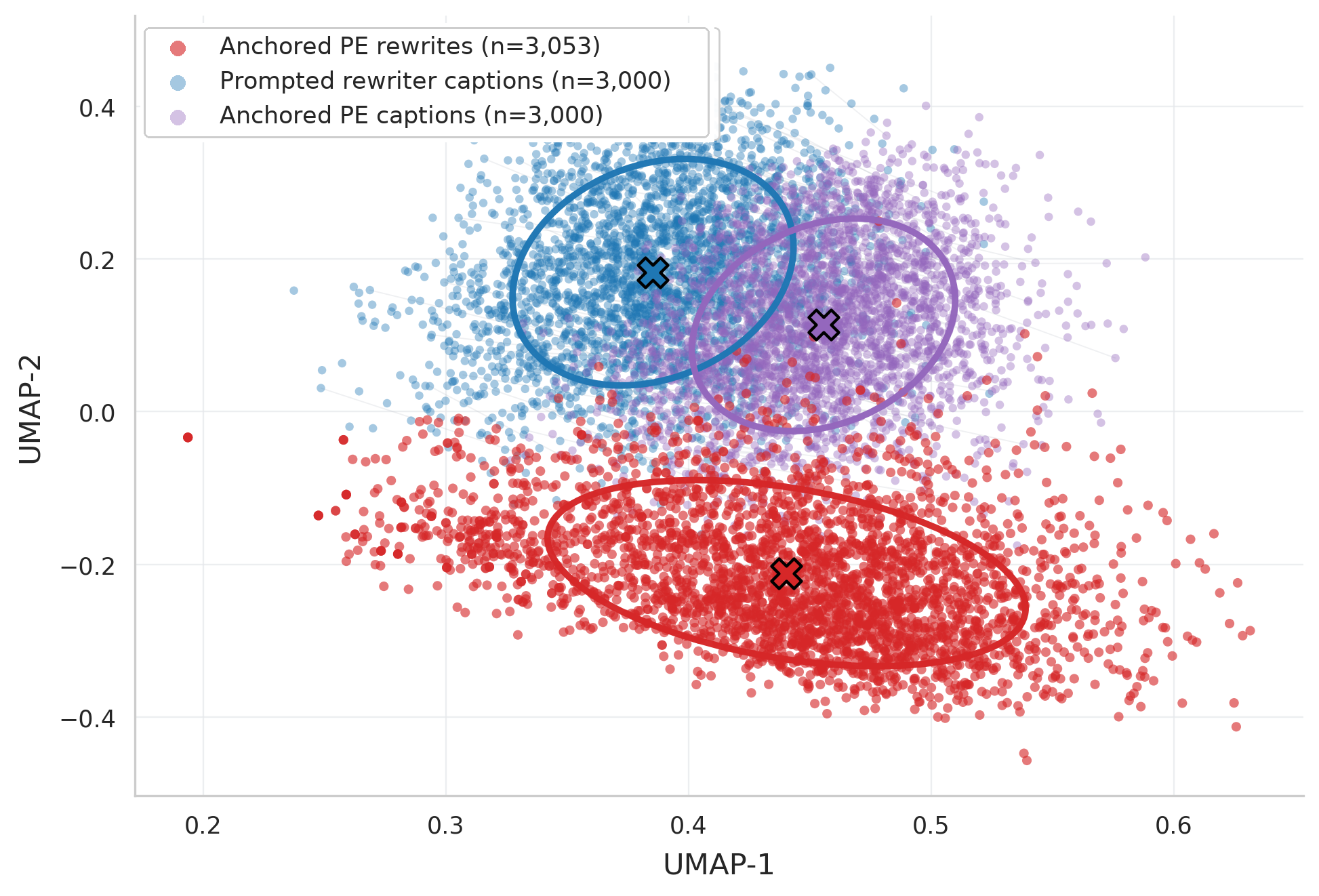}
\caption{Qualitative projection of the three text distributions. The reference set contains one Anchored PE rewrite sampled with seed 42 for each of the 3,053 user prompts pooled across StoryEval, VBench-2.0, and T2V-CompBench. The two fine-tuning-caption sets are sampled from matched dense-caption sources. Crosses mark centroids, and ellipses summarize the projected distributions.}
\label{fig:supp_umap_visual_alignment}
\end{figure}

\paragraph{UMAP visualization settings.}
Figure~\ref{fig:supp_umap_visual_alignment} visualizes the reference PE outputs together with 3,000 paired DiT fine-tuning examples, each providing one prompted rewriter caption and one Anchored PE caption. We jointly represent the reference PE outputs and the 6,000 DiT fine-tuning captions using TF--IDF unigram and bigram features, reduce them to 128 dimensions using truncated SVD, and apply L2 normalization. We then fit one UMAP~\citep{mcinnes2018umap} projection to all three groups using cosine distance, 18 neighbors, a minimum distance of 0.04, and random seed 59 (umap-learn v0.5.11). Thus, all points share the same feature space and projection. This visualization is used only as a qualitative diagnostic.

\subsection{Qwen Embedding Distribution Analysis}

We encode inference-time Anchored PE outputs and DiT fine-tuning captions with frozen Qwen3-VL-Embedding-8B, producing 4096-dimensional L2-normalized vectors. Explicit numbered six-part headers are removed and whitespace is normalized for every condition. The two DiT fine-tuning caption sets are derived from the same dense captions.

For embedding sets $X=\{x_i\}_{i=1}^{m}$ and $Y=\{y_j\}_{j=1}^{n}$, we estimate their discrepancy using the unbiased squared maximum mean discrepancy~\citep{gretton2012kernel}
\[
\begin{aligned}
\widehat{\operatorname{MMD}}_{u}^{2}(X,Y)
={}& \frac{1}{m(m-1)}\sum_{i\ne i'} k(x_i,x_{i'}) \\
&+ \frac{1}{n(n-1)}\sum_{j\ne j'} k(y_j,y_{j'}) \\
&- \frac{2}{mn}\sum_{i=1}^{m}\sum_{j=1}^{n} k(x_i,y_j).
\end{aligned}
\]
where $k(x,y)=\exp\!\left(-\lVert x-y\rVert_2^2/(2\sigma^2)\right)$ is an RBF kernel. Within each analysis, both DiT fine-tuning caption sets use the same bandwidth $\sigma$, selected by the median heuristic. Because this is an unbiased finite-sample estimator of MMD$^2$, its empirical value can be slightly negative; smaller values indicate a smaller estimated discrepancy in the selected embedding space.

The full-data analysis compares the reference PE outputs with both 54K DiT fine-tuning caption sets. The prompted rewriter captions obtain an MMD$^2$ estimate of 0.0758, whereas the Anchored PE captions obtain 0.0673. Their paired difference is $-0.0085$ (95\% CI [$-0.0090,-0.0080$]).

We additionally construct length-matched subsets containing 3,000 reference PE outputs and 3,000 paired examples from each DiT fine-tuning caption set. Matching is based only on Qwen token length and does not imply semantic correspondence between an evaluation-prompt output and a training caption. Matched token counts differ by at most two Qwen tokens. The prompted rewriter captions obtain an MMD$^2$ of 0.0764, while the Anchored PE captions obtain 0.0682. Their difference is $-0.0082$ (95\% CI [$-0.0094,-0.0069$], paired permutation $p{=}0.0005$). The similar reduction after length matching shows that the result is not explained by token length alone.

Confidence intervals use 2,000 paired bootstrap replicates, and the length-matched test uses 2,000 paired label permutations. Pairing preserves the common source-caption IDs in the DiT fine-tuning set when contrasting the two caption sets. These analyses show that Anchored PE captions have a smaller measured embedding discrepancy to the sampling-seed-42 outputs from the Anchored PE; they do not imply that the underlying text distributions are identical or explain the video-score gains.

\subsection{Dense-Reference Preservation and Reformulation Details}

We compare the two DiT fine-tuning caption sets with their shared dense-caption sources over the full 54K training set. Each strictly paired example contains one dense caption, one prompted rewriter caption, and one Anchored PE caption associated with the same training video. We remove numbered six-part field headings, collapse redundant whitespace, and otherwise preserve the complete body text and its order. Table~\ref{tab:supp_dense_reference} reports the paired diagnostics.

We compute ROUGE-L F1 with Porter stemming~\citep{lin2004rouge}. Corpus-level BLEU~\citep{papineni2002bleu} is computed with SacreBLEU 2.6.0~\citep{post2018sacrebleu}; its signature is \path{BLEU|nrefs:1|case:mixed|eff:no|tok:13a|smooth:exp|version:2.6.0}. For paired uncertainty estimates, we additionally compute smoothed sentence-level BLEU-4 for each example and average the resulting scores. Semantic similarity is the cosine similarity between each derived caption and its dense reference encoded by Qwen3-VL-Embedding-8B~\citep{qwen3vlembedding}. Paired 95\% confidence intervals use bootstrap resampling over matched training examples~\citep{koehn2004bootstrap}.

\begin{configtable}
\begin{tabularx}{\linewidth}{@{}p{0.27\linewidth}rrX@{}}
\toprule
\textbf{Metric} & \textbf{Rewrite} & \textbf{Anchored} & \textbf{$\Delta$ (95\% CI)} \\
\midrule
ROUGE-L F1 & 59.79 & 58.03 & $-1.77$ [$-1.84,-1.70$] \\
Mean sentence BLEU-4 & 43.73 & 42.19 & $-1.54$ [$-1.61,-1.47$] \\
Corpus BLEU & 43.90 & 42.13 & $-1.77$ (CI not applicable) \\
Qwen cosine & 0.93467 & 0.93610 & $+0.00143$ [$+0.00112,+0.00173$] \\
\bottomrule
\end{tabularx}
\caption{Dense-reference diagnostics on the paired 54K set. Rewrite denotes the prompted rewriter captions and Anchored denotes the Anchored PE captions. $\Delta$ is Anchored minus Rewrite; intervals are paired 95\% bootstrap CIs.}
\label{tab:supp_dense_reference}
\end{configtable}

Anchored PE captions have lower lexical overlap with the dense references under both ROUGE-L and BLEU, while their mean Qwen embedding cosine is slightly higher. This pattern is consistent with greater lexical and local-sequence reformulation without a corresponding reduction in embedding-level semantic similarity. It does not establish higher caption quality, complete semantic preservation, or a causal link to the downstream video-score gains.

\section{Definition of the Six-Part Caption Schema}

Table~\ref{tab:sixpart_schema} gives the operational definition of the shared
six-part schema used for prompted rewriter outputs, Anchored PE outputs, and
DiT conditioning. The fields provide complementary views of the same
underlying content rather than six independent prompts. At inference, they are
serialized in the order shown below, and the complete six-part text is passed
to the generator.

\begin{table*}[!t]
\centering
\small
\begin{tabularx}{0.96\textwidth}{@{}>{\raggedright\arraybackslash}p{0.18\textwidth}XX@{}}
\toprule
\textbf{Field} & \textbf{Definition and role} & \textbf{Content boundary} \\
\midrule
\textbf{1. Short Caption} &
One concise sentence summarizing the principal subject, action, and scene. It
provides a compact global description of the clip. &
Retains the core event and explicit constraints. For concise-input expansion,
compatible detail may be supplied by the paired captioner-generated target. \\

\textbf{2. Dense Caption} &
A coherent expanded account of the visible content, including subjects,
actions, interactions, scene context, and event order. This is the main
fine-grained semantic description. &
Preserves temporal or logical order and does not contradict explicit input
constraints. Rewriting dense inputs remains conservative; concise-input
expansion may realize compatible detail from the paired target. \\

\textbf{3. Main Object Caption} &
A focused description of the primary person, animal, object, vehicle, or other
salient entity and its principal action, pose, or interaction. &
Includes identity, count, appearance, or attributes only when supported; may be
null when no primary entity can be identified. \\

\textbf{4. Background Caption} &
A description of the environment and surrounding scene, such as location type,
spatial context, weather, lighting, architecture, or nearby elements. &
Contains only grounded setting information; may be null when the source gives
no environmental evidence. \\

\textbf{5. Camera Caption} &
A description of viewpoint, framing, shot scale, and camera behavior, including
static, handheld, pan, tilt, zoom, tracking, aerial, or point-of-view capture. &
Distinguishes camera motion from subject motion. When unspecified, it states
that no particular camera behavior is supported rather than inventing one. \\

\textbf{6. Style Caption} &
A description of visual or media presentation, such as live action, animation,
CGI, game footage, surveillance, screen recording, slow motion, or time lapse. &
Avoids subjective quality terms (e.g., ``cinematic'' or ``beautiful'') unless
grounded. When unspecified, it records the absence of a supported special
style. \\
\bottomrule
\end{tabularx}
\normalsize
\caption{Operational definition of the six-part caption schema. Dense-caption rewriting remains grounded in the source caption, video-grounded captioning remains grounded in visible evidence, and concise-input training supervises expansion toward its paired captioner-generated target without contradicting the input.}
\label{tab:sixpart_schema}
\end{table*}

The schema decomposes a video description into six complementary levels. The
\emph{Short Caption} supplies a compact global summary of the principal subject,
action, and scene, whereas the \emph{Dense Caption} expands this summary into the
main semantic account, including interactions, event progression, and relevant
context. The remaining fields isolate factors useful for video generation.
The \emph{Main Object Caption} focuses on the salient entity and its
attributes or behavior; the \emph{Background Caption} describes the environment
and surrounding elements; the \emph{Camera Caption} records viewpoint, framing,
and camera motion; and the \emph{Style Caption} specifies the visual or media
presentation. This separation makes content, cinematography, and style explicit
while keeping them mutually consistent.

The schema does not license unsupported or contradictory detail. For
informative caption inputs, rewriting preserves claims stated or strongly
implied by the source. For concise training inputs, the paired
captioner-generated targets may contain compatible visual detail while
preserving the explicit input constraints. During video-grounded captioning,
claims must be supported by visible evidence. Missing object or background
information may be represented as null.
For camera and style, an explicit statement that no particular property is
supported is preferred to inventing a plausible one. Temporal order,
uncertainty, names, visible text, and other source-specific information are
preserved whenever present. In CAPE-T2V, both the prompted rewriter and the
Anchored PE produce captions in this same ordered structure. The six fields are
then serialized into one prose conditioning sequence and supplied in full to
the DiT. The shared schema fixes the field structure; comparing outputs from
the two caption operators then isolates differences in organization, level of
detail, wording, and overall output distribution.

% Let the wide schema table share a page with the explanatory text above, then
% settle it before the next section begins.  Placing this barrier immediately
% after table* creates a vertically centered float-only page.
\FloatBarrier
\raggedbottom
\section{Representative Training Pairs for PE Anchoring}

The SFT mixture used for PE anchoring contains inputs with substantially
different information densities, all paired with targets in a structurally
consistent six-part format. The examples below give complete targets from the
three input constructions. The pseudo user prompt and detailed source caption
correspond to the same clip, whereas the concise source caption comes from a
separate example.

\begin{trainingcase}{Concise source caption input}
\caseinput{A person pouring food into a pot on a stove.}

\begin{targetcase}{sfttargetcolor}
\begin{sixpartcaption}
\sixpartfield{1. Short Caption}{A person carefully pours a mixture of chunked
vegetables and pieces of meat or tofu from a pink bowl into a black pot
containing vigorously boiling water on a stove.}
\sixpartfield{2. Dense Caption}{In a domestic kitchen setting, a person stands
before a stove where a black pot rests on an active burner. The pot holds water
that is boiling vigorously, releasing thick clouds of steam and forming large,
rolling bubbles on the surface. The individual tilts a bright pink bowl, guiding
a colorful assortment of chunked vegetables, such as carrots and onions, along
with cubes of meat or tofu, into the turbulent liquid. As the ingredients slide
from the bowl, they splash gently into the hot water, integrating with the
boiling broth while the steam rises around the cooking vessel. The scene
captures the dynamic motion of the food transfer and the intense heat of the
cooking process.}
\sixpartfield{3. Main Object Caption}{The primary subject is a person holding a
pink bowl and pouring its contents, a mix of chunked vegetables and meat or tofu
pieces, into a black pot.}
\sixpartfield{4. Background Caption}{The background features a kitchen
environment centered on a stove with an active burner supporting the black pot
and rising steam.}
\sixpartfield{5. Camera Caption}{A clear wide view with centered composition
keeps the described subjects and actions visible.}
\sixpartfield{6. Style Caption}{The visual presentation stays focused on the
described subjects and actions.}
\end{sixpartcaption}
\end{targetcase}

\casebehavior{The target expands the core cooking action into all six fields: it
summarizes a person pouring ingredients into a boiling pot, identifies the
person and pot as the main subjects, places the action in a kitchen, and uses
conservative camera and style descriptions. Fine details in this target
originate from the upstream detailed caption paired with the brief input,
rather than being inferred from the concise text alone.}
\end{trainingcase}

\begin{trainingcase}{Pseudo user prompt input}
\caseinput{Close-up of hands in blue gloves making a pie crust on a stainless
steel counter, crimping edges and rolling dough, bright instructional kitchen
video.}

\begin{targetcase}{sfttargetcolor}
\begin{sixpartcaption}
\sixpartfield{1. Short Caption}{A bright instructional kitchen video presents a
close-up of hands in blue gloves making a pie crust on a stainless steel counter
by rolling dough and carefully crimping the edges.}
\sixpartfield{2. Dense Caption}{This bright instructional kitchen video features
a tight close-up of hands wearing blue disposable gloves working diligently on a
stainless steel counter. The sequence begins with the hands rolling out smooth,
pale dough into an even circle using a wooden rolling pin. Next, the baker lifts
the dough into a pie dish and meticulously crimps the edges to form a
decorative, fluted crust. The clean metal surface reflects the overhead
lighting, emphasizing the texture of the flour-dusted dough and the precise
finger movements involved in shaping the pastry. Every step highlights the
technique required for perfect crust formation in a professional culinary
setting, showcasing the tactile process of baking from start to finish with
clear visibility.}
\sixpartfield{3. Main Object Caption}{The primary subjects are a pair of hands
clad in blue gloves actively rolling dough and crimping the edges of a pie crust
on a work surface.}
\sixpartfield{4. Background Caption}{The setting is a bright, professional
kitchen environment centered around a clean, reflective stainless steel counter
that supports the baking activity.}
\sixpartfield{5. Camera Caption}{The camera maintains a close-up view to keep the
hands and pie crust clearly visible.}
\sixpartfield{6. Style Caption}{The visual presentation stays focused on the
described subjects and actions.}
\end{sixpartcaption}
\end{targetcase}

\casebehavior{The captioner-generated target preserves the blue gloves, pie
crust, stainless-steel counter, rolling and crimping actions, close-up framing,
and instructional-video style. It expands these constraints into a coherent
preparation sequence while retaining the user's principal intent.}
\end{trainingcase}

\begin{trainingcase}{Detailed source caption input}
\caseinput{In a bright, professional kitchen with a clean stainless steel
countertop, a person wearing blue gloves meticulously crafts a pie crust. The
stationary camera maintains a close-up focus on the hands as they shape the
dough into a tin, crimp the edges for a decorative finish, and roll it to an
even thickness. Finally, the baker uses a fork to prick holes in the bottom of
the crust, demonstrating precise technique in this clear, instructional view of
the pie-making process.}

\begin{targetcase}{sfttargetcolor}
\begin{sixpartcaption}
\sixpartfield{1. Short Caption}{In a bright professional kitchen, a person
wearing blue gloves meticulously crafts a pie crust on a stainless steel
countertop by shaping dough into a tin, crimping the edges, rolling it to an even
thickness, and finally pricking holes in the bottom with a fork.}
\sixpartfield{2. Dense Caption}{In a bright, professional kitchen featuring a
clean stainless steel countertop, a person wearing blue gloves meticulously
crafts a pie crust. The hands shape the dough into a tin, crimp the edges to
create a decorative finish, and roll the material to an even thickness. Finally,
the baker uses a fork to prick holes in the bottom of the crust. This sequence
demonstrates precise technique throughout the clear, instructional view of the
pie-making process, focusing entirely on the manual steps involved in preparing
the pastry within the specified workspace setting.}
\sixpartfield{3. Main Object Caption}{The main subject is a person wearing blue
gloves who shapes dough into a pie tin, crimps the edges, rolls the dough to an
even thickness, and pricks holes in the bottom with a fork.}
\sixpartfield{4. Background Caption}{The background consists of a bright,
professional kitchen featuring a clean stainless steel countertop that serves as
the workspace.}
\sixpartfield{5. Camera Caption}{A stationary camera maintains a close-up focus
on the hands and the pie crust, keeping the described actions clearly visible.}
\sixpartfield{6. Style Caption}{The visual presentation stays focused on the
described subjects and actions.}
\end{sixpartcaption}
\end{targetcase}

\casebehavior{The target faithfully reorganizes the full event chain across the
six fields, including shaping, crimping, rolling, and pricking the dough,
together with the explicit stationary close-up camera cue and kitchen setting.
Unlike the concise-input construction, this pair mainly teaches structure
preservation rather than controlled completion from an underspecified input.}
\end{trainingcase}

\clearpage
\flushbottom
\section{Representative DiT Training Pair}

\begin{figure*}[!t]
\centering
\includegraphics[width=0.98\textwidth]{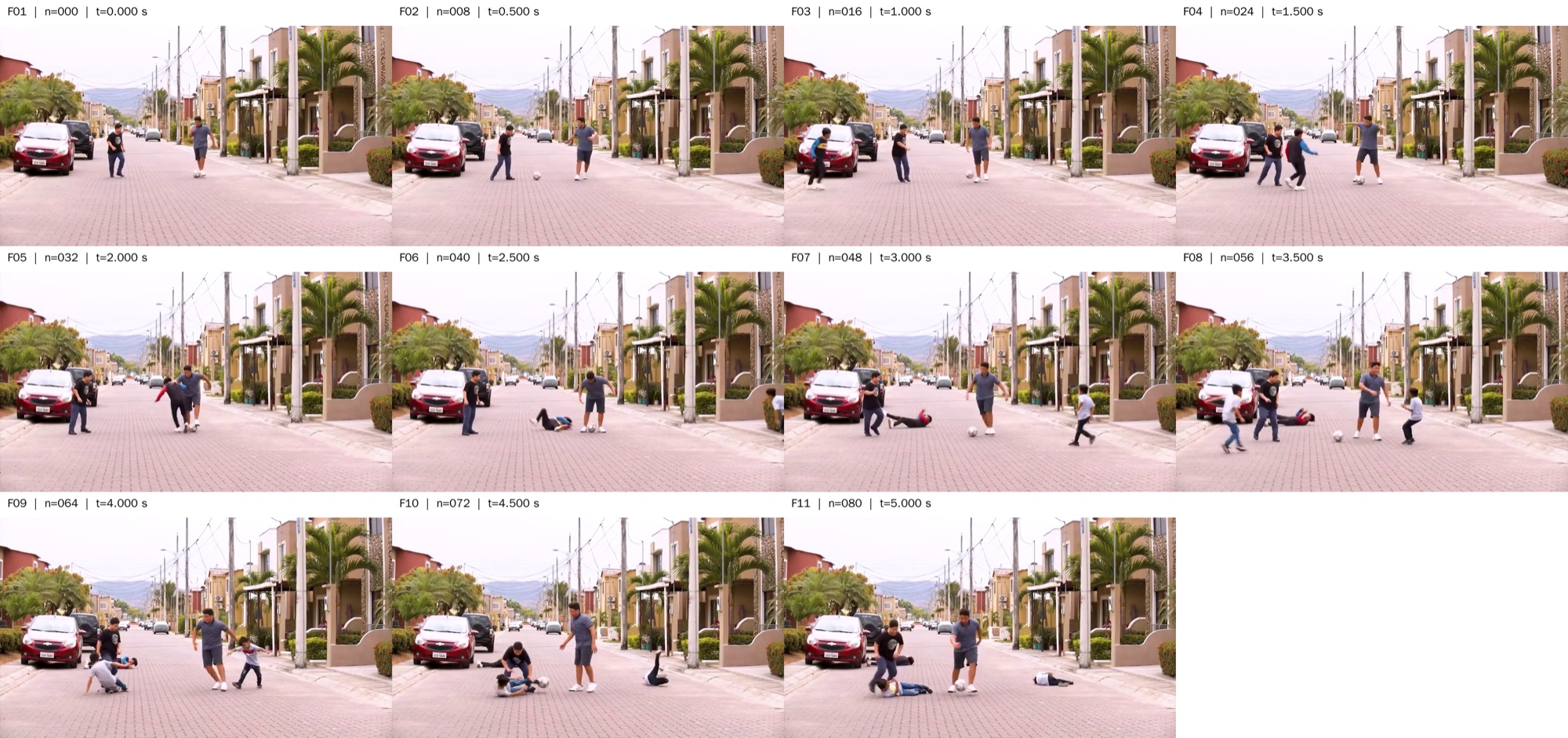}
\caption{Representative DiT fine-tuning video. Eleven frames sampled at 2 fps show the multi-event sequence; labels give zero-based source-frame indices and timestamps.}
\label{fig:dit_training_case}

\vspace{4pt}
\begin{minipage}{0.96\textwidth}
\centering
\small
\begin{tabularx}{\linewidth}{@{}p{0.13\textwidth}p{0.25\textwidth}p{0.25\textwidth}X@{}}
\toprule
\textbf{Field} & \textbf{Prompted rewriter caption} & \textbf{Anchored PE caption} &
\textbf{Observed change} \\
\midrule
Short Caption &
\emph{``A man ... dribbles ... while three boys attempt to
tackle him and sequentially fall''} &
\emph{``In a wide, static shot ... three boys approach and fall
after attempting to challenge him''} &
Moves camera and scene context into the global summary; compresses the precise
``sequentially'' wording into a more general event statement. \\

Dense Caption &
Introduces the \emph{wide static shot} before describing each
challenger and concludes that all three lie on the street. &
Keeps the scene description in the body and ends with
\emph{``leaving all three challengers on the ground while the man
remains standing with the ball.''} &
Removes duplicated camera wording from the dense field and tightens the
terminal-state summary while preserving event order. \\

Main Object &
Enumerates the boys' \emph{black, blue/red, and white attire}. &
Groups them as three boys who
\emph{``approach from the left and right, attempt to challenge him,
and fall''}. &
Trades clothing detail for entry direction and a shared action trajectory. \\

Background &
\emph{``A paved residential street lined with beige and red
houses and parked cars.''} &
\emph{``The background features''} the same street, houses, and
cars. &
Content is unchanged; only the field-oriented phrasing differs. \\

Camera &
\emph{``Wide, static shot.''} &
\emph{``The camera maintains a wide, static shot throughout the
sequence.''} &
Converts a label-like fragment into an explicit temporal statement. \\

Style &
\emph{``Realistic live-action footage.''} &
\emph{``The visual presentation stays focused on the described
subjects and actions.''} &
Uses a more conservative formulation but drops the explicit live-action label. \\
\bottomrule
\end{tabularx}
\normalsize
\captionof{table}{Field-level comparison of matched prompted rewriter and Anchored PE captions. Contrasting excerpts identify changed content; the final column describes the redistribution without implying a quality ranking.}
\label{tab:dit_training_case_diff}
\end{minipage}
\end{figure*}

Figure~\ref{fig:dit_training_case} shows a representative video retained in the
54K-video DiT fine-tuning set. The 5.06-second clip contains a multi-part event:
a man controls a soccer ball on a residential street while three boys approach
from different directions, attempt to challenge him, and fall in sequence. The
prompted rewriter and Anchored PE use the same video and dense-caption source;
only the caption operator differs. This paired example therefore illustrates
the change in conditioning text while holding the video content fixed;
Table~\ref{tab:dit_training_case_diff} summarizes the field-level differences.

\begin{captioncase}{Prompted rewriter caption}{rewritercolor}
\textbf{1. Short Caption:} A man in a grey t-shirt dribbles a soccer ball on a
residential street while three boys attempt to tackle him and sequentially fall
to the ground.

\textbf{2. Dense Caption:} In a wide, static shot of a paved residential street
lined with beige and red houses and parked cars, a man wearing a grey t-shirt
and dark shorts dribbles a white soccer ball. He is approached by a boy in a
black t-shirt and jeans, while a second boy in a blue and red long-sleeved shirt
runs in from the left to challenge for the ball but collides with the man and
falls. As the man continues to control the ball, a third boy in a white t-shirt
enters from the right. The boy in black attempts a tackle and falls, followed
immediately by the boy in white sliding on the pavement and falling as well. The
scene ends with all three challengers lying on the street while the man in grey
remains standing with the ball.

\textbf{3. Main Object Caption:} A man in a grey t-shirt and dark shorts
dribbling a white soccer ball, surrounded by three boys in varying attire (black
t-shirt and jeans, blue and red long-sleeved shirt, white t-shirt) who attempt
to tackle him and fall.

\textbf{4. Background Caption:} A paved residential street lined with beige and
red houses and parked cars.

\textbf{5. Camera Caption:} Wide, static shot.

\textbf{6. Style Caption:} Realistic live-action footage.
\end{captioncase}

\begin{captioncase}{Anchored PE caption}{anchoredcolor}
\textbf{1. Short Caption:} In a wide, static shot of a paved residential street,
a man in a grey t-shirt dribbles a white soccer ball while three boys approach
and fall after attempting to challenge him.

\textbf{2. Dense Caption:} On a paved residential street lined with beige and
red houses and parked cars, a man in a grey t-shirt and dark shorts dribbles a
white soccer ball. A boy in a black t-shirt and jeans approaches him, followed
by a second boy in a blue and red long-sleeved shirt running from the left who
collides with the man and falls. As the man continues to control the ball, a
third boy in a white t-shirt enters from the right. The boy in black attempts a
tackle and falls, immediately followed by the boy in white sliding on the
pavement and falling, leaving all three challengers on the ground while the man
remains standing with the ball.

\textbf{3. Main Object Caption:} The main subjects are a man in a grey t-shirt
dribbling a white soccer ball and three boys who approach from the left and
right, attempt to challenge him, and fall to the ground.

\textbf{4. Background Caption:} The background features a paved residential
street lined with beige and red houses and parked cars.

\textbf{5. Camera Caption:} The camera maintains a wide, static shot throughout
the sequence.

\textbf{6. Style Caption:} The visual presentation stays focused on the
described subjects and actions.
\end{captioncase}

Both captions are complete outputs under the same six-part schema. The Anchored
PE caption moves the camera specification into the global summary and groups the
entry directions, ordered falls, and terminal state across the Dense and Main
Object fields. The case is illustrative rather than evidence that one caption
is universally superior; its purpose is to make the controlled difference in
caption operator concrete.

\section{Prompt Availability}
\label{sec:supp_prompt_availability}

The complete prompts used for Step~1 pair construction and consistency filtering, Anchored
PE inference, Step~2 video-grounded dense-caption construction, and
video--caption consistency filtering are provided in the project repository at
\url{https://github.com/yizzz927/CAPE-T2V}. They are omitted here to avoid
duplicating implementation details. The repository also includes the fixed
six-part instruction used by the prompted rewriter.

% Full-width qualitative figures are intentionally last so that their float
% placement cannot interrupt any subsequent appendix text.
\FloatBarrier
\raggedbottom
\suppshowcases

\FloatBarrier
\end{document}